\documentclass[%
 reprint,
 amsmath,amssymb,
 aps,
]{revtex4-2}

\usepackage{graphicx}
\usepackage{dcolumn}
\usepackage{bm}
\usepackage{xcolor}
\usepackage[T2A]{fontenc}

\usepackage{amsmath}
\usepackage{amssymb}
\usepackage{mathtools}
\usepackage{amsthm}
\usepackage{algorithm}
\usepackage{algpseudocode}
\usepackage{hyperref}
\algnewcommand\algorithmicforeach{\textbf{for each}}
\algdef{S}[FOR]{ForEach}[1]{\algorithmicforeach\ #1\ \algorithmicdo}

\begin{document}

\preprint{AIP/123-QED}

\title[SNN Global Review]{Contemporary implementations of spiking bio-inspired neural networks}


\author{Andrey E. Schegolev}
 \email{a.e.schegolev@pn.sinp.msu.ru.}
 \affiliation{Skobeltsyn Institute of Nuclear Physics, Lomonosov Moscow State University, 119991 Moscow, Russia}
 
\author{Marina V. Bastrakova}
\author{Michael A. Sergeev}
\affiliation{Faculty of Physics, Lobachevsky State University of Nizhni Novgorod,  603950 Nizhny Novgorod, Russia
}

\author{Anastasia A. Maksimovskaya}
 \affiliation{Faculty of Physics, Lomonosov Moscow State University, 119991 Moscow, Russia
 }
\affiliation{Dukhov All-Russia Research Institute of Automatics, Moscow 101000, Russia}

\author{Nikolay V. Klenov}
 \affiliation{Faculty of Physics, Lomonosov Moscow State University, 119991 Moscow, Russia
}
\affiliation{Dukhov All-Russia Research Institute of Automatics, Moscow 101000, Russia}

\author{Igor I. Soloviev}
 \affiliation{Skobeltsyn Institute of Nuclear Physics, Lomonosov Moscow State University, 119991 Moscow, Russia
}
\affiliation{Dukhov All-Russia Research Institute of Automatics, Moscow 101000, Russia}

\date{\today}

\begin{abstract}
The extensive development of the field of spiking neural networks has led to many areas of research that have a direct impact on people's lives. As the most bio-similar of all neural networks, spiking neural networks not only allow the solution of recognition and clustering problems (including dynamics), but also contribute to the growing knowledge of the human nervous system. Our analysis has shown that the hardware implementation is of great importance, since the specifics of the physical processes in the network cells affect their ability to simulate the neural activity of living neural tissue, the efficiency of certain stages of information processing, storage and transmission. This survey reviews existing hardware neuromorphic implementations of bio-inspired spiking networks in the "semiconductor", "superconductor" and "optical" domains. Special attention is given to the possibility of effective "hybrids" of different approaches
\end{abstract}


\maketitle

\section{Introduction}  


The last decade has demonstrated a significant increase in interdisciplinary research in neuroscience and neurobiology (this was reflected even in the decisions of the Nobel Committee in 2024 \cite{abramson2024accurate, Hopfied1982}). 
The convergence of mathematics, physics, biology, neuroscience and computer science has led to the hardware realisation of numerous models that mimic the behaviour of living nervous tissue and reproduce characteristic neural patterns. Spiking Neural Networks (SNNs) have played a crucial role in these fields of knowledge. In these networks, neurons exchange short (about 1...2 ms for bio-systems) pulses of the same amplitude (about $100~mV$ for bio-systems) \cite{prieto2016neural, taherkhani2020review}. SNNs come closest to mimick the activity of living nervous tissue (capable of solving surprisingly complex tasks with limiting resources) and have the greatest biosimilarity and bioinspirability.

Spiking neural networks use a completely distinct method of information transfer between neurons: they encode input data as a series of discrete time spikes that resemble the action potential of biological neurons. In fact, the fundamental idea of SNNs is to achieve the closest possible bio-similarity and use it to solve specific tasks. These problems 
can be roughly divided into two groups: the first group is focused on solving traditional neural network challenges, with more emphasis on dynamic information recognition (speech, video), while the second group is aimed at imitating the nervous activity of living beings, reproducing characteristic activity patterns, recreating the work of the human brain. Currently, there is an ambitious project that aims to create a full-fledged artificial mouse brain \cite{markram2006blue}. Moreover, the second group includes such tasks as: using motor biorhythms for neural control in robotics \cite{van2016neurorobotics, iosa2016three, falotico2017connecting}, controlling human movements (bioprosthetics, functional restoration of mobility) \cite{raspopovic2021neurorobotics}, understanding learning processes and memory effects \cite{kasabov2014neucube, yamazaki2022spiking, liang2020temporal}, creating brain-computer interfaces \cite{he2020brain, gao2021interface, yadav2020comprehensive, flesher2021brain, bonci2021introductory}, etc.

Therefore, hardware development of bio-inspired SNN is very vital and promising. For this reason, it is crucial to recognise the current progress and conditions for the development of this field, taking into account the immense amount of information that is growing every day. Moreover, in terms of signalling, SNN is better suited to hardware implementation than artificial neural networks (ANN), since neurons are only active at the time when a voltage spike is generated, which reduces the overall power consumption of the network and simplifies computation.

\begin{figure}
    \centering
    \includegraphics[width=0.9\linewidth]{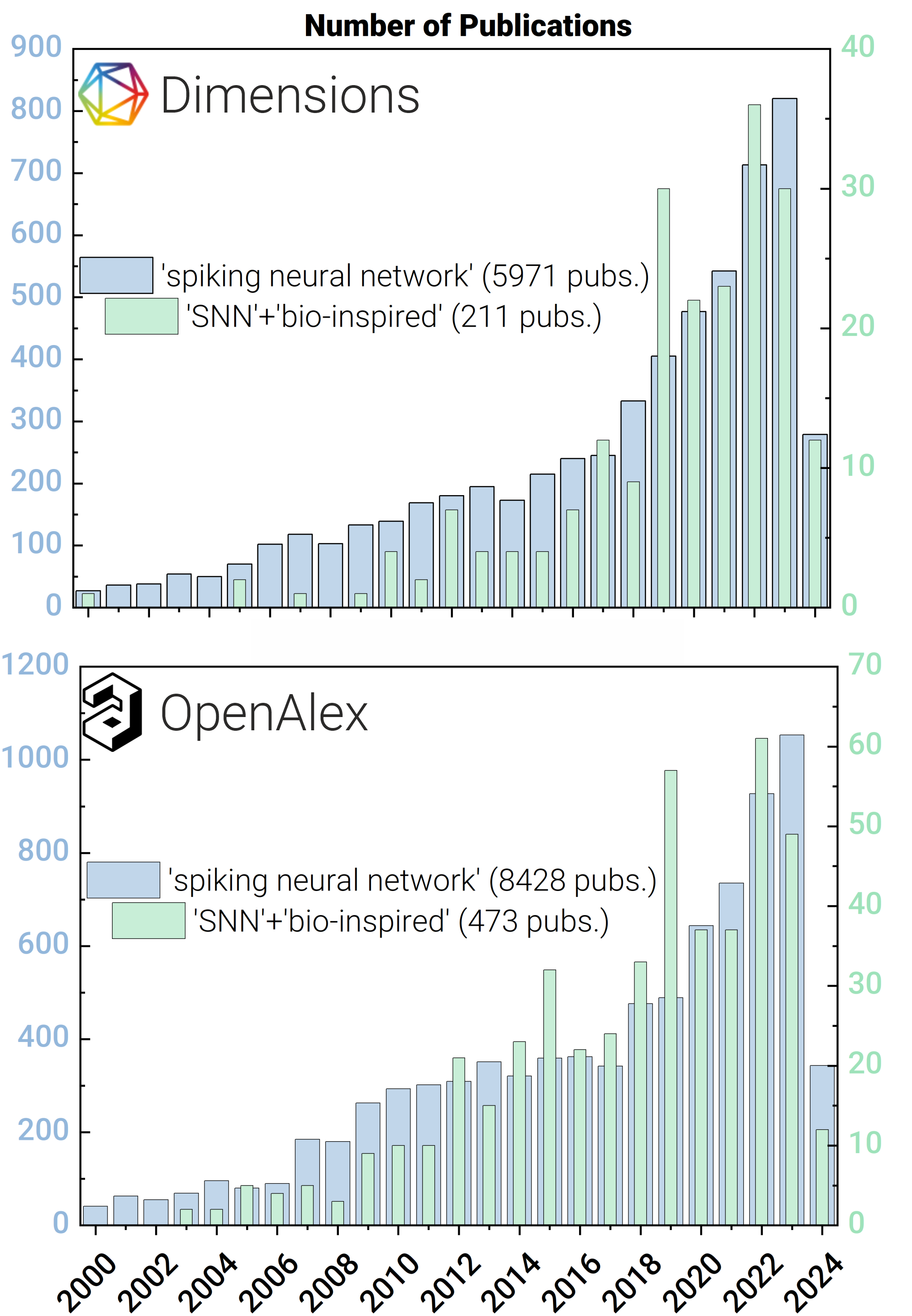}
    \caption{The number of publications histograms based on data from Dimensions and OpenAlex databases from a search on the topic ``spiking neural network'' (light blue histogram) and a search filtered by the keyword ``bio-inspired'' (light green histogram), presented retrospectively from the year 2000.}
    \label{fig:infographic}
\end{figure}

Figure \ref{fig:infographic} illustrates the intensification of work on the topic over the last decades. This is the period with a significant increase in publication activity. Here we show the analysis of publication activity (indexed in Dimensions and OpenAlex databases) on the topic of spiking neural networks and on the topic of spiking neural networks filtered by the keywords ``bio-inspired''. Since around 2016, there has been a significant increase in interest in the topic of spiking neural networks as well as in the topic of bio-inspired networks. All studies presented in these publications have been carried out for different implementations of SNNs: software, complementary metal–oxide–semiconductor, or, shortly, CMOS (especially memristor-based), superconducting, optical, and hybrid ones.

In this paper, we review the advances in different hardware directions of bio-inspired spiking neural network evolution and provide a brief summary of the future of the field, its advantages, challenges, and drawbacks. Furthermore
we have reviewed the basic software implementations of SNNs. 

The first part of this review is devoted to the mathematical foundations of bio-inspired spiking neural networks, their idea and software realisation. The second part is dedicated to the CMOS-based bio-inspired neuromorphic circuits, where we have talked about semiconducting and memristive realisations. The third part is devoted to superconducting realisations, not of the whole network, but of its elements and some of its functional parts. Finally, we have considered bio-inspired elements for optical neuromorphic systems and provide a brief discussion of each area of activity as well as a general conclusion on the evolution of bio-inspired neuromorphic systems.

The brief conclusion is that there is no single approach has overwhelming advantages at the current moment, and it is quite likely that the necessary direction of development of the field has not yet been found. However, we can already say with certainty that the hybrid approach can provide some success in the formation of complex deep spiking neuromorphic networks.

\section{Mathematical fundamentals of bio-inspired spiking neural networks}

\begin{figure}
    \centering
    \includegraphics[width=0.7\linewidth]{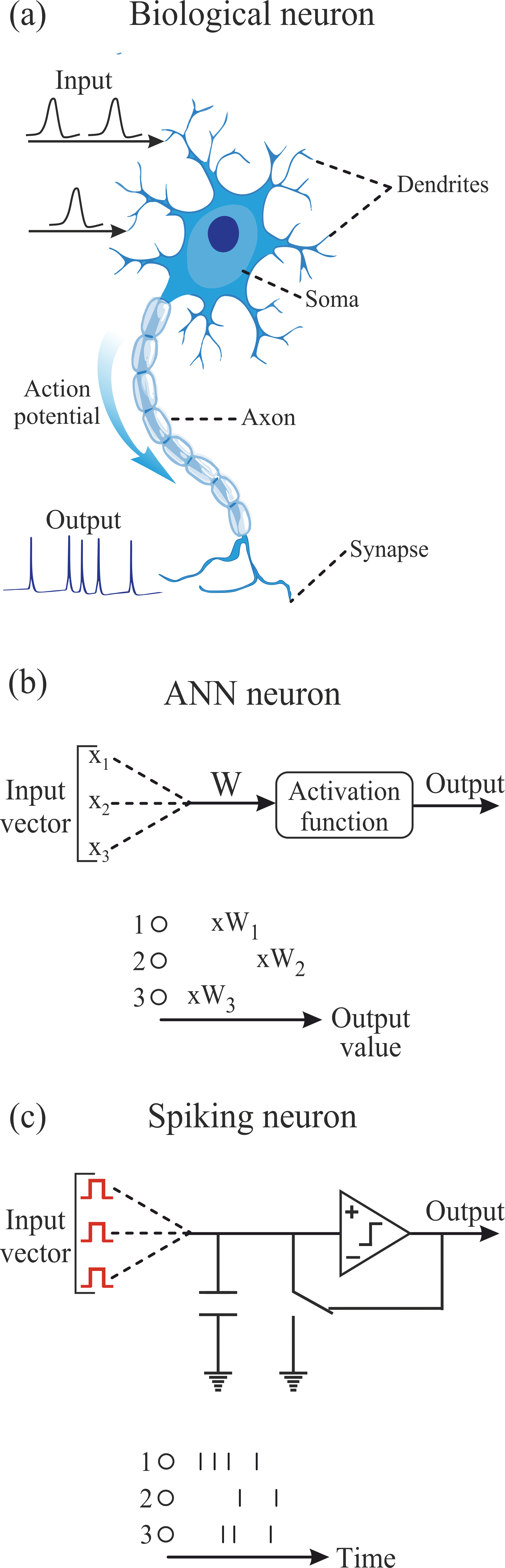}
    \caption{\cite{zheng2020operational} a - Illustration of a biological neuron and its operating principles.  b - Conventional ANN  operating principle (top) and output representation (bottom). The input vector is processed by multiplying the input vector by the corresponding weight vector (denoted $W$) and then passing it to the activation function. Output values are represented in floating-point numbers that can be processed at the software level. c - Operating principle of a spiking neural network (top) and output representation (bottom). The input signal is processed by the hardware implementation of the neuron. Output values are represented in spike trains, which differ in the emission time of each spike and the overall density, and also serve as inherent memory. 
    }
    \label{fig:SNN basic}
\end{figure}

A spiking neural network is fundamentally different from the second generation of neural networks: instead of continuously changing values over time, such a network works with discrete events (chains of events) that occur at specific points in time. Discrete events are encoded by impulses (spikes) received at the input of the neural network and processed by it in a specific way, as shown in figure \ref{fig:SNN basic}. The output of such a network is also a sequence of spikes, encoding the result of the network's activity. In a real neuron, the transmission of impulses is described by differential equations that correspond to the physical and chemical processes of action potential formation. When the action potential reaches its threshold, the neuron generates a spike and the membrane potential returns to its initial level, see figure \ref{fig:SNN basic}(a). An accurate representation of neuronal activity and its response to various input signals requires a general mathematical model that describes all the necessary processes associated with its spike activity and action potential formation, while remaining sufficiently simple for its successful use in various applications.

\subsection{Mathematical models of bio-inspired neurons and networks}
\subsubsection{Hodgkin-Huxley model}

The discussion of applications begins with the simplest and most studied software implementations of SNNs: the most popular way to describe the initiation and propagation of action potentials in neurons is the Hodgkin-Huxley (HH) model \textcolor{red}{\cite{hodgkin1952quantitative}}.

The HH model is treated as a conductance-based system where each neuron is a circuit of parallel capacitors and resistors \cite{malcom2023comprehensive} and describes how action potential in nerve cells (neurons) is emerged and propagated. Basically, HH model contains \cite{hausser2000hodgkin} four components of the current flowing through the neuron membrane, formed by lipid bilayer and possessing of potential $V_m$: the current flowing through the lipid bilayer ($I_c$), currents flowing through the ion channels ($I_{Na}$ and $I_{K}$), and the leakage current ($I_l$). The lipid bilayer -- nerve cell membrane -- is introduced in the form of the capacitance $C_m$, ion-gated channel -- by the conductance $g_i$ (where $i$ is a corresponding ion) per unit area and leakage current introduced by the conductance $g_l$ per unit area. Thus, for a cell model that contains only sodium and potassium ion channels, the total current flowing through the membrane is described by the following equation
\begin{equation} \label{eq:hh}
    I = C_m \frac{dV_m}{dt} + g_K(V_m - V_K) + g_{Na}(V_m - V_{Na}) + g_l(V_m - V_l).
\end{equation}


Of course, the biochemical processes in the living nerve cell are more complicated and therefore the HH model could be modified by adding extra terms for other ions ($Cl^-$ or $Ca^{++}$, for example) or by using non-linear conductance models ($g_i = f(t, V)$) instead of constant values. 

\begin{figure}[h!]
    \centering
    \includegraphics[width=0.75\linewidth]{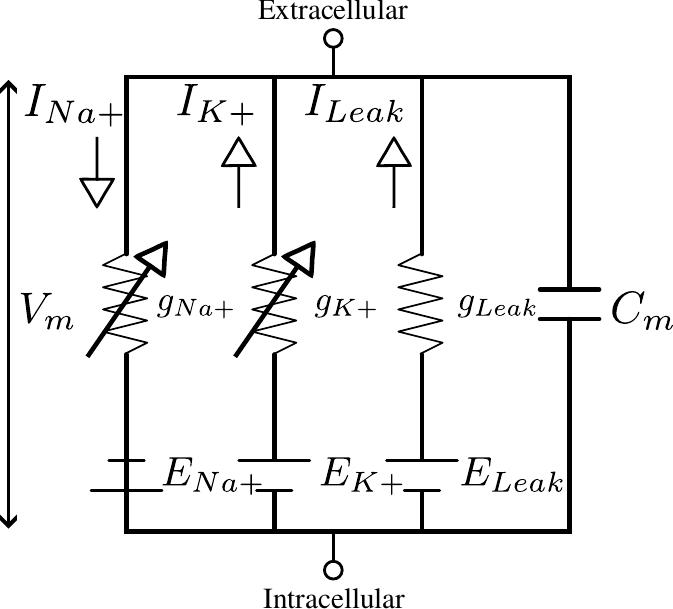}
    \caption{\cite{phdthesis} Electrical analog of the Hodgkin-Huxley model. The capacitance $C_m$ represents a lipid bilayer -- neuron membrane with potential $V_m$. Non-linear electrical conductances resemble voltage-gated ion channels $g_{K+}$ and $g_{Na+}$, while $g_{Leak}$ corresponds to the leakage channel. Parameters $E_{Na+}$, $E_{K+}$, and $E_{Leak}$ correspond to the reversal ions potentials and leak reversal potential, respectively.
    }
    \label{fig:HH_model}
\end{figure}

\subsubsection{Izhikevich model}

One of the most computation-efficient and simultaneously accurately representative for neurons' activity is the Izhikevich model. According to Ref. \cite{Izhikevich}, bifurcation methodologies \cite{Izhikevich_bifurcations} enable us to reduce many biophysically accurate Hodgkin–Huxley-type neuronal models to a two-dimensional (2-D) system of ordinary differential equations
\begin{equation}\label{Izch_if}
\begin{cases}
    \dot{v} = 0.04v^2 + 5v + 104 - u + I, \\
    \dot{u} = a(bv-u),
\end{cases}
\end{equation}
with the auxiliary after-spike resetting
\begin{equation}\label{Izc_if_2}
    \text{if} \quad v\geq 30 mV, \quad \text{then} 
    \begin{cases}
    & v \leftarrow c, \\
    & u \leftarrow u+ d.
    \end{cases}   
\end{equation}

Here, $v$ and $u$ are variables, $a,b,c,d$ are parameters, and dot notation is the time derivative. Neuron's membrane potential and recovery are described by $v$ and $u$, and when $v$ reaches its maximum value of $+30mV$, $u$ is restored, see Eq. \eqref{Izc_if_2}. They are affected by the parameter $b$ which describes recovery variable sensitivity to sub-threshold fluctuations of $v$, promoting possible low-threshold oscillations when increased. The case $b < a$ $(b > a)$ corresponds to saddle-node (Andronov–Hopf) bifurcation of the resting state.
Variable $u$ is also affected by the parameter $a$ which stands for the time scale of the variable, speeding up or slowing down recovery, and the parameter $d$, which is responsible for after-spike reset of $u$ caused by slow high-threshold $Na^+$ and $K^+$ conductances. Fast high-threshold $K^+$ conductances affect on $v$ is expressed in the remaining parameter $c$, that describes the after-spike reset value of $v$. 

The combination $0.04v^2+5v+140$ provides scaling for membrane potential $v$ to $mV$ and $ms$ scale and was derived from fitting the initiation dynamics of the spike. The resting potential in the model is between $-70$ and $-60$ $mV$ depending on the $b$. But in real practice, the model does not have a fixed threshold, depending from the value of $v$ before the spike formation.

\subsubsection{Leaky Integrate-and-Fire Neuron Model}

The really frequently used hardware model is implemented on principles of so-called \textit{Leaky Integrate-and-Fire} (LIF) neurons. Such neurons, whose membrane potentials $\mathbf{V}$ evolve at time step $t$ according to some variation of the following dynamics, described in following manner \cite{wu2018spatio}:
\begin{equation}
    \mathbf{V}[t] = \lambda \mathbf{V}[t-1] + \mathbf{WX}[t] - \mathbf{S}[t]V_{th},
\end{equation}
where $\lambda \in [0, 1)$ is a membrane leakage constant, $\mathbf{X}$ is an input (i.e., an external stimuli to the network or the spiking activity from another neuron), $\mathbf{W}$ is a weight matrix, and the binary spiking function
\begin{equation}
    \mathbf{S}[t] = \begin{cases} 
    1, & \text{if $\mathbf{V}[t] > V_{th}$} \\ 0, & \text{otherwise}
    \end{cases}
\end{equation}
is a function of the threshold voltage $V_{th}$.


\subsection{Coding information in SNN} 


The brain of a living being processes a wide variety of information, which can come from different sensory organs as well as signals from the nervous system of different internal organs. How does the brain understand which signal comes from the optic nerve and which from the auditory nerve, for example? There are different ways of encoding incoming information, processing it by a specially trained neural network (a specific area in the brain) and interpreting the results of the processing. In other words, there are ways of encoding input and output signals. And both are very important: in the first case, the information is encoded so that the system can work with it, and in the second case, so that the general decision-making centre (the brain) can perceive or $handle$ the processed information. Two types of coding are generally used in artificial SNNs: rate coding and temporal (or latency) coding, although there are others \cite{eshraghian2023training}.

Rate coding converts input information intensity into a ``firing'' rate or spike count. For example, a pixel in some image, that has a specific RGB code and brightness, can be convert or $coding$ into a Poisson train -- a sequence of spikes, based on this information (information of its intensity). There are several types of rate coding: count rate coding, density rate coding and population rate coding. In terms of output interpretation, the processing centre will select the one that has the highest ``firing'' rate or spike count at a particular point in time. 

Another natural principle of information coding is a temporal (or latency) coding. One converts input intensity to a spike time, referring to spike timing and paying attention to time moment when the spike has occurred. The spike weighting ensures that different ``firing'' times lead to different amount of information. The earlier a spike arrives, the larger weight it carries, and the more information it transmits to the post-synaptic neurons. In terms of output interpretation, the processing centre selects the signal that, by a certain point in time, came first (selects the signal from that output neuron among other output neurons that fused first).

There is also another type of information coding called delta modulation. This type of coding consists of converting the incoming analogue information signal into a spike train of temporal changes in signal intensity (magnitude). For example, if the input signal is increasing, the network input will receive spikes at a certain frequency, which depends on the rate of increase of the signal (its derivative) -- the faster the signal increases in time, the more frequent the spikes will be. Conversely, the signal decreasing will be accompanied by the absence of spikes.

\subsection{SNN learning techniques}

The complex dynamics of spike propagation over SNNs makes it difficult to design a learning algorithm that gives the best result. Currently, there are three main types of methods for training spiking neural networks, as referred in Ref. \cite{eshraghian2023training}: shadow training, backpropagation on spikes training and training based on local learning rules.


\subsubsection{Shadow training}


The idea of this shadow training technique is to use the training algorithms of a conventional ANN to build a spiking network by converting a trained ANN into a trained SNN. This process takes place as follows: first, the conventional ANN is trained, then the activation function of each neuron in the ANN is replaced by a separate operator that non-linearly transforms the signal incoming to the neuron by the spike frequency or by the delay between spikes. For example, conversion process from convolutional neural network to SNN can be done by manually reprogramming convolution kernel for spike train inputs in order to make the SNN produce the same output as the trained convolutional neural network \cite{perez2013mapping}. At the same time, the net weights remain the same.



The main advantage of this training process is that the most of the time we deal with an ANN, with all of its benefits of training conventional neural networks. Such approach is used in tasks related to that of ANNs that aimed to image classification. Despite that, the process of converting activation functions into spike trains usually requires a large number of simulations time steps which may deteriorate the initial idea of spiking neuron energy efficiency. And, the most important one, is that conversion process is doing an approximation of activation, negatively affecting the performance of a SNN. The last further confirms the fact that other learning algorithms need to be developed to train SNNs, till then SNNs will remain just a shadow of ANNs.

\subsubsection{Backpropagation training}

Backpropagation training algorithm for ANNs can also be implemented for SNNs by calculating gradients of weight change for every neuron. Implementations of this algorithm for SNNs may vary since, according to original ANNs backpropagation the learning requires to calculate the derivative of the loss function. And as we know, the uses of derivative over spikes is not the best idea, because spikes generation depends on membrane threshold potential as a step-function and the derivative becomes infinite. To avoid this, backpropagation techniques for SNN take other neuron output signal parameters into account, i.e. backpropagation method using spikes utilises changes of spike timing rate according to the network weights change.

Backpropagation method has several advantages, such as the high performance on data-driven tasks, low energy consumption, and high degree of similarity with recurrent neural networks in terms of training process. Despite the similarity with well-established ANN backpropagation method, the drawbacks of backpropagation for SNNs method consist of several subjects. First of all, this method can not fully replicate effectiveness of optimising a loss function, meaning that there is still an accuracy gap between SNNs and ANNs, which remains to be closed up for today. Additionally, once neurons become inactive during the training process, their weights become frozen.

In some cases, there are alternative interpretations of this algorithm, for example, such as the Forward Propagation Through Time (FPTT) \cite{yin2023accurate} which is used for recurrent SNN training. This algorithm is devoid of many drawbacks that follow conventional backpropagation algorithm, removing the dependence on partial gradients sum during the gradient calculation. The most peculiar feature of this algorithm is that along with regular loss it computes dynamic regularisation penalty, which is calculated on previously encountered loss value, transforming recurrent nertwork training to resemble feed-forward network training.

\subsubsection{Local learning rules}

Neurons are trained locally, treating local spatially and temporary signal as an input for a single neuron's weight update function. Namely, backpropagation technique deals only with finite sequences of data, restricting the temporal dependencies that can be learned. This algorithm also tries to compute gradients as it is done in backpropagation, but does it through the computations that make these gradient calculations temporally local. However, this algorithm demands significantly more computations in comparison with backpropagation, which rejects the possibility of this algorithm to replace conventional backpropagation despite being more biologically plausible.

The constraint imposed on brain-inspired learning algorithms is that the calculation of a gradient should, like the forward pass, be temporally local, i.e. that they only depend on values available at either present time. To address this, learning algorithms turn to their online counterparts that adhere to temporal locality. Real-time recurrent learning (RTRL) \cite{irie2023exploring} proposed back in 1989 is one prominent example. 

\subsection{Towards hardware implementations of SNN}
To extend the understanding of the concept of how spiking neural networks work, in this section we will review a schematic process of solving the exclusive OR (XOR) problem with a SNN on an example presented in Ref. \cite{reljan2017solving}. Since XOR is considered a fairly simple logic gate, it will serve as an excellent demonstration of SNN training and operation.
In fact, operating with SNNs has almost the same challenges as in training ANNs: for example, which data representation to choose to effectively describe the problem, what structure should the network have, how to interpret the network output and so on. That is, we should guide our approach to a certain problem in the same way as it is done in ANNs on the general scope, including the XOR problem.

The training process for the chosen task can be separated into several steps:
\begin{itemize}
    \item Neuron model \\
    Due to low complexity of this task, it is most efficient to use a simple neuron model as well. In that case, the LIF model suits this problem the best because its easy to control and does not produce unwanted complexity.
    
    \item Data representation \\
    Since we test the possibility of creating a network of spiking neurons to perform logic gate operations, the input data can be simply labeled in two inputs. Each input has to carry its own representative frequency in order to be separable for the network. Taking these statements into account, we can (for example) define the frequency input values as 25 Hz for the Boolean Zero (``0'') and 51 Hz for the Boolean One (``1''). These values are chosen as they are decisively different from each other and can be clearly represented visually.
    
    \item Structure \\
    The Neural network structure represents the logic gate as follows: two input neurons that encode any combination of ``0'' and ``1'', hidden layer neurons represent any possible combination of 0's and 1's, and output layer produces the predicted output, as shown in figure~\ref{fig:SNN structure experim}.
    
    \begin{figure}[h!]
    \centering
    \includegraphics[width=0.6\linewidth]{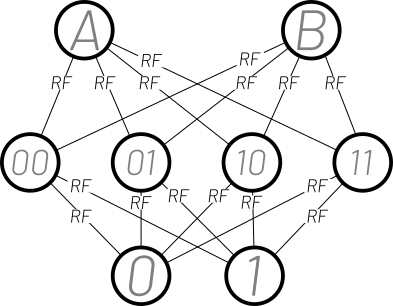}
    \caption{\cite{reljan2017solving}SNN architecture for the XOR problem solving. Input neurons $A$ and $B$ can receive any combination of ``0'' and ``1''. Each $RF$ block demonstrates Receptive Field application to neuron input/output. The number on hidden neurons layer represent the combination of ``0'' and ``1'' they implied to respond to.
    }
    \label{fig:SNN structure experim}
    \end{figure}
    
    \item RFF \\
    To strengthen the response for designated frequencies for the corresponding neurons, Receptive Fields filter (RFF) is added to every hidden layer neurons' connections. Adding the RFF helps neurons to detect particular frequencies, filtering the ones that are outside their response range. The RFF formula is $k_{ij}={e^{-((x_m-y_0)/d_m)^2}}$ where $k_{ij}$ is a scalar variable which will modify the output spike train frequency of the related neuron, $x_m$ is the operating frequency of the RFF, $y_0$ is the input spike train frequency to the RFF and $d_m$ denotes the width of the RFF, allowing the filter to distinguish more frequencies.
    \item Input encoding \\
    The input frequencies are encoded into linear spike trains, i.e. the value of the distance between the action potentials, known as the inter spike interval (ISI), is treated to be a constant. The network was designed to take advantage of the precise timing between action potentials. If the ISIs on input $A$ are synchronised with the ISIs on input $B$, it means that both inputs have identical frequency.
    
    \item Training \\
    Usage of RFF is an effective and biologically plausible way to reduce complexity and fault vulnerability in this problem. Because of that, as well as small number of neurons, manual fine-tuning of thresholds and weights will be enough to train the network to produce accurate results, successfully recognising all possible XOR combinations. However, upon increasing the size of the network, implementing one of network training algorithms will be inevitable.

\end{itemize}

For conclusion to this example it is important to highlight that one should pay attention to training process workflow and investigate the problem deeply to be able to recognise all the obstacles that may be encountered during training (i.e. in the XOR problem proper signal frequency interpretation would not be possible without RFF). Considering the training workflow, the most important parts of this are the neuron model and input/output coding, since it can determine crucial aspects of problem interpretation from network implementation's point of view. Conversely, network training algorithms for weights optimisation play lesser role in this process, because they can only affect computational complexity to be performed and their influence on results accuracy becomes noticeable only if the task implies implementing state-of-the-art network or results close to it.


\section{CMOS-based bio-inspired neuromorphic circuits}
\begin{figure}[h!]
    \centering
    \includegraphics[width=0.95\linewidth]{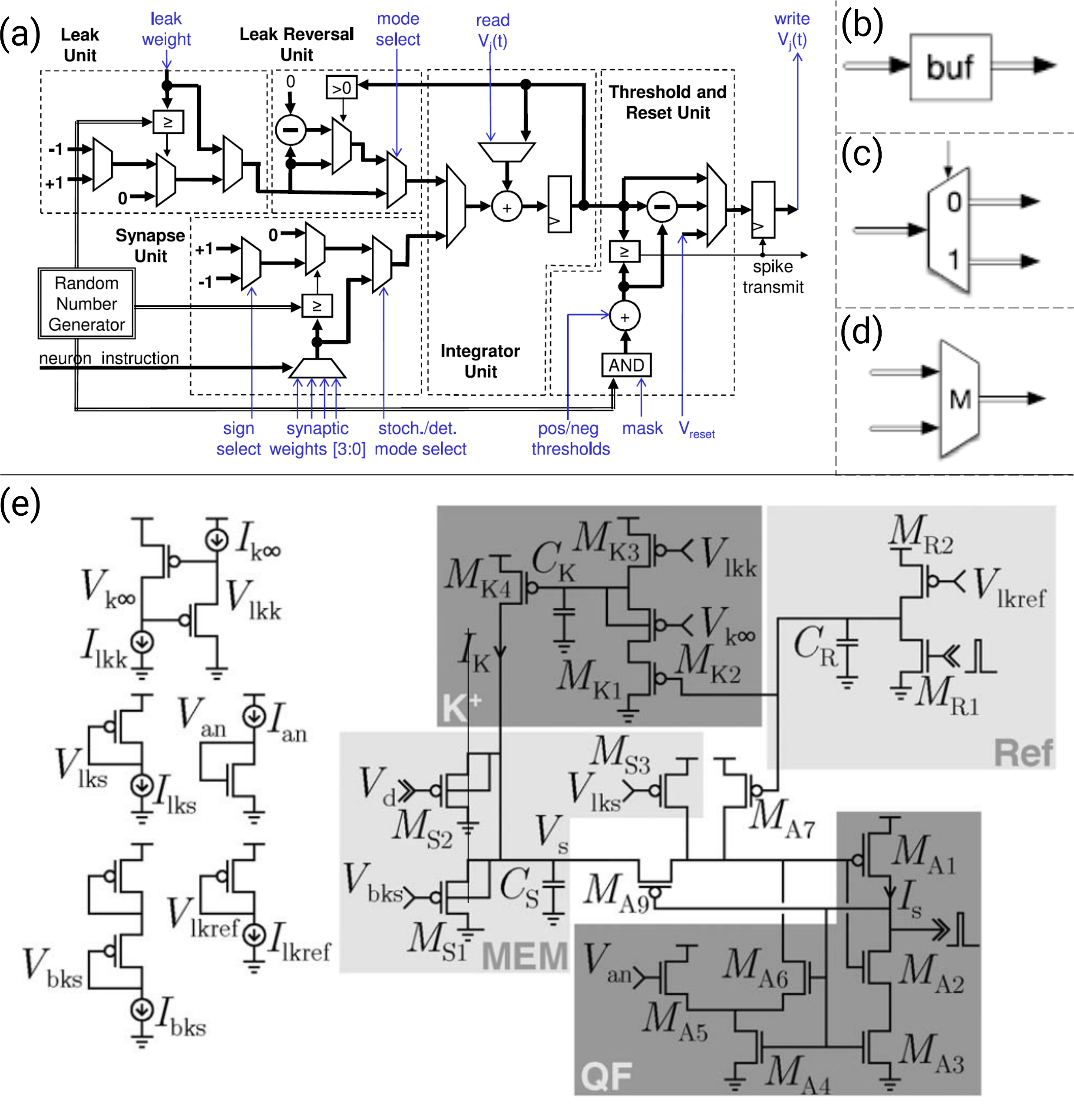}
    \caption{Semiconductor solutions for realising bio-inspired neuron functions: (a) TrueNorth neuron circuit \cite{akopyan2015truenorth}, (b) buffer gate, (c) controlled split gate, (d) non-deterministic merge gate, (e) Neurogrid neuron's soma circuit \cite{benjamin2014neurogrid}
    }
    \label{fig:CMOS_neurons}
\end{figure}

The last decades of electronics and electrical engineering are clearly associated with the development of the conventional complementary metal-oxide-semiconductor (CMOS) technology \cite{indiveri2011neuromorphic, merolla2014million}. Semiconductors are well established in many devices that we use 7 days a week, all year round. It is therefore not surprising to see variations in the hardware implementation of SNNs based on semiconductor elements. Globally, all CMOS SNN implementation options can be divided into two parts, with the focus on either transistors or memristors. Examples of semiconductor implementations of neuron circuits are shown in figure~\ref{fig:CMOS_neurons}. Two CMOS solutions for implementing the functions of the single neuron model developed by IBM (for TrueNorth) and Stanford University (for Neurogrid) are presented here. In both cases, the realisation of the functions of even a single neuron (more precisely, the neuron soma) requires the involvement of a large number of elements.

\subsection{Silicon-based neuron: operation}

As might be expected, the realisation of spike sequence generation in CMOS technology differs from the way it is done in software models. Since semiconductor circuits operate with voltage levels, generating a spike as a non-linear voltage response requires some tricks. 
To better understand how SNNs work at the hardware level and how spike formation occurs, let us briefly review the operation of the basic elements of such networks, which are isomorphic to ion-gated channels \cite{indiveri2011neuromorphic, camunas2019neuromorphic}.

\subsubsection{Realisation of ion-gated mechanisms}

The goal of creating semiconductor models of neuromorphic spiking neural networks is to reproduce the biochemical dynamics of ionic processes in living cells. The first step is to reproduce the ion-gated channels responsible for the different voltage response patterns. The Hodgkin-Huxley (HH) neuron model discussed above is essentially a thermodynamic model of ion channels. The channel model consists of a number of independent gating particles that can adopt two states (open or closed) which determine the permeability of the channel. The HH variable represents the probability that a particle is in the open state or, in population terms, the fraction of particles in the open state.

In a steady state, the total number of opening particles (the opening flux, which depends on the number of closed channels and the opening rate) is balanced by the total number of closing particles (the closing flux). A change in the membrane voltage (or potential) causes an increase in the rate of one of the transitions, which in turn causes an increase in the corresponding flux of particles, thereby modifying the overall state of the system. The system will then reach a new steady state with a new ratio of open to closed channels, thus ensuring that the fluxes are once again in equilibrium.

This situation becomes even clearer if we consider it from the perspective of energy balance. Indeed, a change in the voltage on the membrane is equivalent to a change in the electric field across it. Then the equilibrium state of the system depends on the energy difference between the particles in the different states: if this difference is zero and the transition rates are the same, the particles are equally distributed between the two states. Otherwise, the state with lower energy will be preferred and the system will tend to move to it. Note that in the HH model, the change in the population of energy states is exponential in time. A similar situation is with the density of charge carriers at the source and drain of the transistor channel, the value of which also depends exponentially on the size of the energy barriers. These energy barriers exist due to the inherent potential difference (electric field) between the channel and the source or drain. Varying the source or drain voltage changes the energy level of the charge carriers.

\begin{figure}
    \centering
    \includegraphics[width=0.95\linewidth]{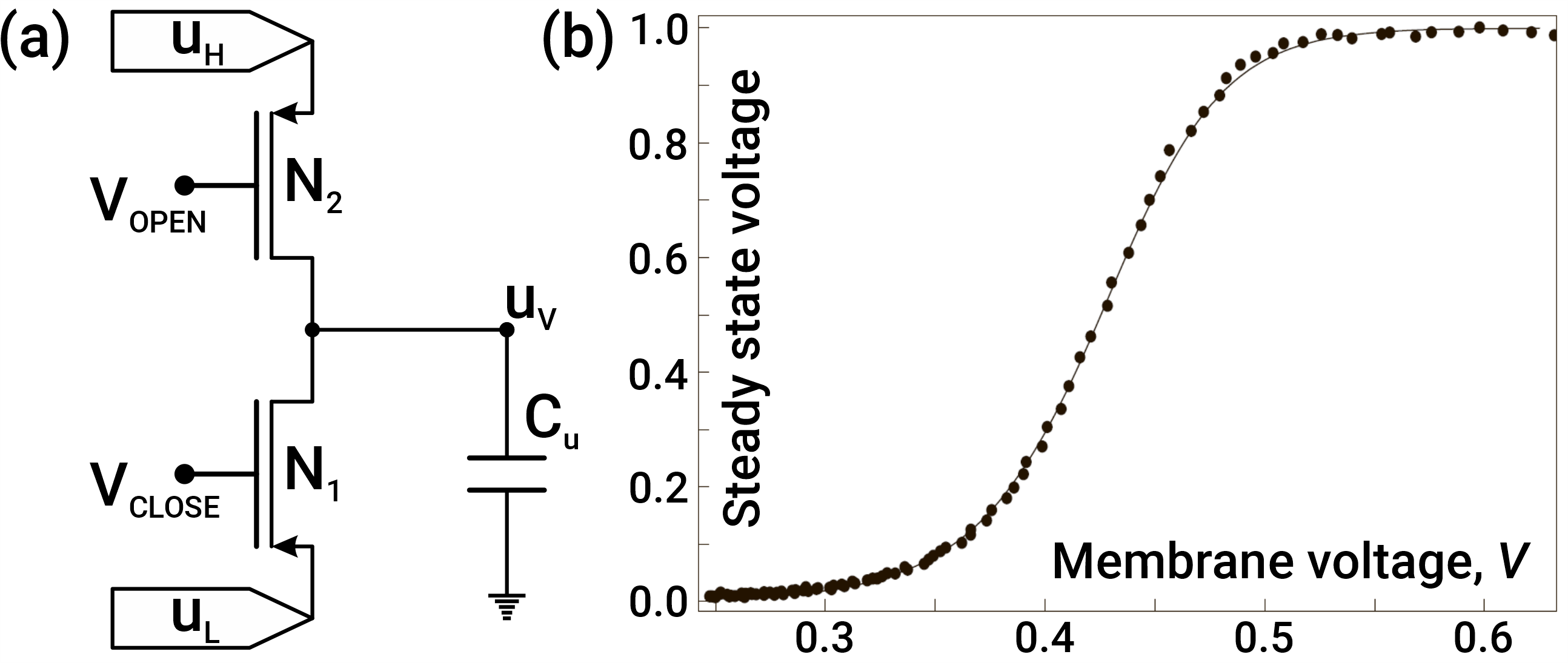}
    \caption{(a) Voltage-Dependent Silicon Ion Channel Model based on two transistors \cite{hynna2007thermodynamically} and  (b) simulation of steady state voltage level ($u_V$) versus the membrane voltage ($V$). Note that $V_{OPEN}$ and $V_{CLOSE}$ are proportional to $V$ and $-V$ respectively.
    }
    \label{fig:ion_channel_model}
\end{figure}

The similarity of the physics underlying the operation of neuronal ion channels and transistors allows us to use transistors as thermodynamic imitators of ion channels. In both cases, there is a direct relationship between the energy barrier and the control voltage. In an ion channel, isolated charges must overcome the electric field generated by the voltage across the membrane. For a transistor, electrons or holes must overcome the electric field created by the voltage difference between the source, or drain, and the transistor channel. In both of these cases, the charge transfer across the energy barrier is governed by the Boltzmann distribution, resulting in an exponential voltage dependence \cite{hynna2007thermodynamically}. To model the closing and opening fluxes, we need to use at least two transistors, the difference of signals from which must be integrated to model the state of the system as a whole. A capacitor ($C_u$) does the integration, assuming that the charge represents the number of particles and the current represents the flux. The voltage on this capacitor is linearly proportional to its charge gives the result. K. Hynna and K. Boahen (2007) suggested the following circuit (figure~\ref{fig:ion_channel_model}, to simulate a voltage-driven ion channel. Each transistor defines an energy barrier for one of the transition rates: transistor $N1$ uses its source and gate voltages ($u_L$ and $V_{CLOSE}$, respectively) to define the closing rate, and transistor $N2$ uses its drain and gate voltages ($u_H$ and $V_{OPEN}$) to define the opening rate (where $u_H > u_L$). The difference in transistors currents is collected on the capacitor $C_u$. The magnitude of the energy barriers is independent of the capacitor voltage $u_V$, so the value of $u_V$ indicates the fraction of open channels and increases as the particles switch more and more to the open state. The variable’s steady-state value changes sigmoidally with membrane voltage, dictated by the ratio of the opening and closing rates.

\subsubsection{Integrate-and-fire neuron realisation}

\begin{figure}
    \centering
    \includegraphics[width=0.95\linewidth]{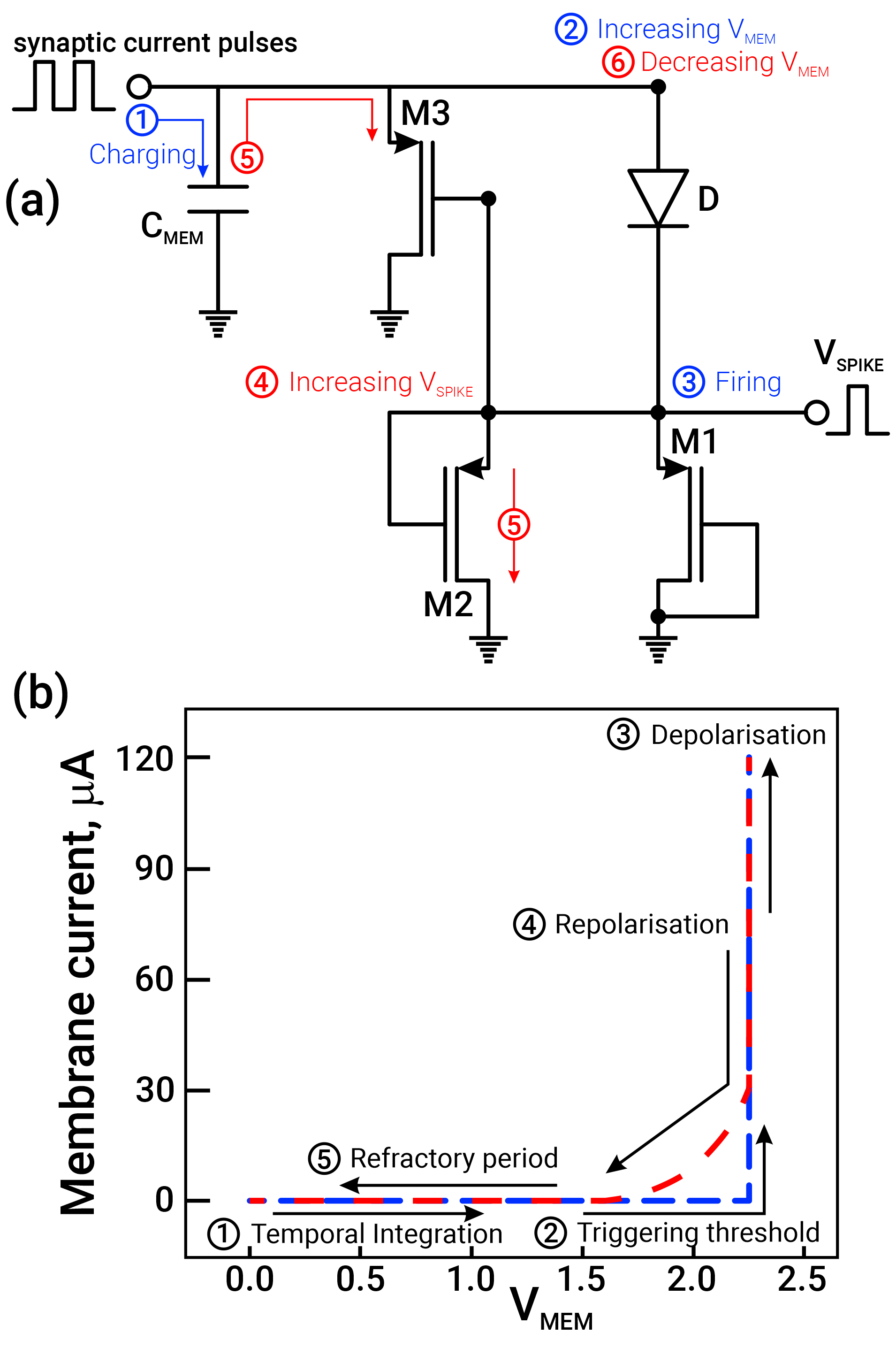}
    \caption{(a) Integrate-and-fire neuron circuit with  the illustration of the operation mechanism and (b) the dependence of membrane current on membrane voltage \cite{park2021integrate}
    }.
    \label{fig:I&F_neuron}
\end{figure}

Now we are ready to understand the membrane voltage and spike formation. The explanation of these processes will be conducted based on a model of an integrate-and-fire neuron realised using MOSFETs ($M1$, $M2$, $M3$), capacitance ($C_{mem}$) and $p-n-p-n$ diode ($D0$) \cite{park2021integrate}, shown in figure~\ref{fig:I&F_neuron}a. The designated components perform the following roles: $C_{mem}$ contributes to the increase in the membrane voltage ($V_{mem}$); $D0$ generates spike voltages ($V_{SPIKE}$); $M1$ acts as a resistor and the resistance contributes to the determination of the $V_{SPIKE}$ value; $M2$ and $M3$ are responsible for resetting the spiking and membrane voltages, respectively.

The presented neuron circuit operation begins with the flow of synaptic current pulses (step 1) from pre-synaptic devices into the neuron circuit. Charges carried by the input current are integrated into $C_{mem}$. The temporal integration of charges increases $V_{mem}$ which is the anode voltage of $D0$ (step 2). $V_{SPIKE}$ is abruptly generated when $V_{mem}$ reaches a triggering threshold voltage for the latch-up of the anode current of the diode (step 3). The $V_{SPIKE}$ value is determined by the voltage division of the diode and $M1$. The generation of $V_{SPIKE}$ supplies the gate voltages to $M2$ and $M3$ and opens the channels of these transistors (step 4). The discharge current flows from $C_{mem}$ to the ground through the M3 channel (step 5), and this flow rapidly decreases $V_{mem}$ (step 6). Simultaneously, the reset current flows from the cathode of $D0$ to the ground through the $M2$ channel (step 5). Eventually, the opening of the $M2$ and $M3$ channels resets the anode and cathode voltages to zero (step 6), and accordingly $V_{SPIKE}$ becomes zero. Thus, the latch-up of $D0$ and the subsequent opening of the $M2$ and $M3$ channels cause the presented neuron circuit to fire $V_{SPIKE}$ pulses toward post-synaptic devices \cite{park2021integrate}.

Figure \ref{fig:I&F_neuron}b illustrates the manner in which the membrane current changes with the membrane voltage during the charging and discharging of $C_{mem}$. We can see that the neuron circuit mimics the (1) temporal integration, (2) trigger threshold, (3) depolarisation, (4) repolarisation and (5) refractory period of a biological neuron. The membrane current does not flow during the temporal integration of charges in $C_{mem}$ (step 1 in figure~\ref{fig:I&F_neuron}). When the temporal integration reaches the threshold voltage (steps 2 and 4), the membrane current increases abruptly (step 3). This moment corresponds to the depolarisation of the electrical action potential in a biological neuron. Discharging $C_{mem}$ after depolarisation (step 5) results in a rapid and then gradual decrease in membrane current (step 6), which corresponds to the repolarisation of the electrical action potential in a biological neuron. As the membrane current becomes negligible, the presented neuron circuit remains in the refractory period \cite{park2021integrate}.

\subsubsection{Izhikevich neuron model with silicon-based realisation}

\begin{figure}
    \centering
    \includegraphics[width=0.95\linewidth]{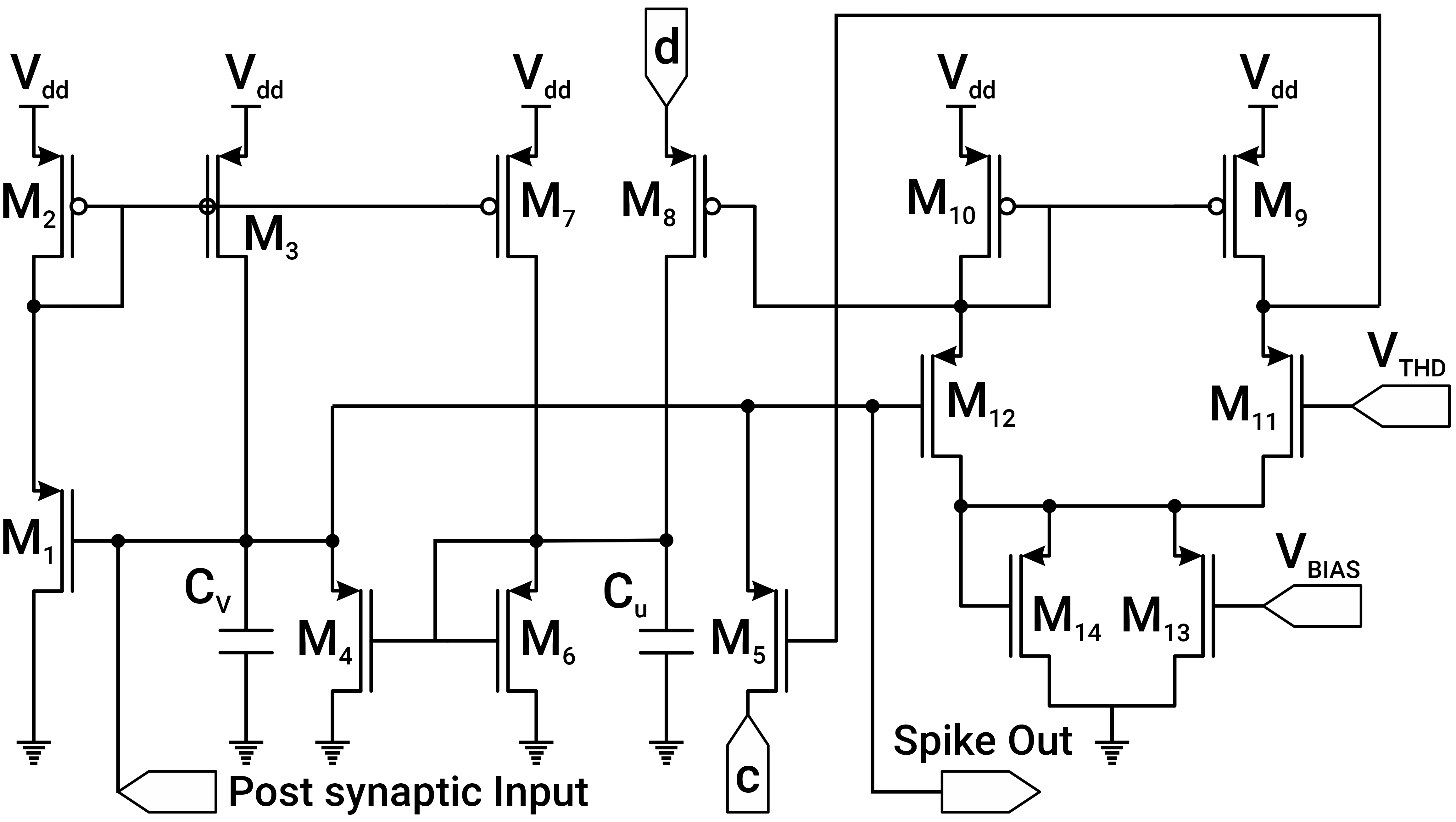}
    \caption{Illustration of the cortical neuron circuit, based on 14 MOSFETs and emulated Izhikevich neuron model \cite{wijekoon2008compact}
    }
    \label{fig:cortical_neuron}
\end{figure}

In 2008 Jayawan H.B. Wijekoon and Piotr Dudek suggested the circuit that implements the cortical neuron (figure~\ref{fig:cortical_neuron}) \cite{wijekoon2008compact}, inspired by the mathematical neuron model proposed by Izhikevich in 2003. The circuit contains 14 MOSFETs, based on which three blocks are operating: membrane potential circuitry (transistors $M1$ -– $M5$ and the capacitance $C_v$), slow variable circuitry (transistors $M1$, $M2$ and $M6$ -- $M8$) and comparator circuitry ($M9$ -- $M14$).

$C_v$ integrated a positive feedback current that generate spikes (generated by $M3$ and flowing from one to $C_v$), leakage current (generated by $M4$) and post-synaptic input current (excitatory or inhibitory). If a spike is generated, it is detected by the comparator circuit which provides a reset pulse on the gate of $M5$ that rapidly hyperpolarises the membrane potential ($V$ on $C_v$) to a value determined by the voltage at node $c$. The transistor $M5$ size is designed so that the capacitor $C_v$ is fully discharge during the voltage pulse coming from the comparator circuit by feedback loop (to the gate of $M5$).

The slow variable circuit works as follows. 
The magnitude of the current supplied by $M7$ is determined by the membrane potential (voltage at $C_v$), in a manner analogous to that observed in the membrane circuit. The scaling of transistors M3 and M7 ensures that the drain current of M7 is less than that of M3, while the capacitance value of $C_u$ is selected to be greater than that of $C_v$. This guarantees that the potential at $C_u$ varies more slowly than at $C_v$. 
The sum of these currents is integrated across the slow variable capacitor, designated as $C_u$. Additionally, following a membrane potential spike, the comparator (right part of figure~\ref{fig:cortical_neuron}) generates a voltage pulse (arriving at the base of $M8$) that opens the transistor, identified as $M8$. 
The modest dimensions of M8 and the brief duration of the voltage pulse that opens it guarantee that the capacitance $C_u$ is not entirely reset to $V_(dd)$. Instead, an additional quantity of charge, regulated by $V_(dd)$, is transferred to $C_u$. 
It can be demonstrated that each membrane spike results in a rapid increase in the slow variable potential. This, in turn, gives rise to an increase in the leakage current of the membrane potential, which in turn causes a slowing down of the depolarisation following the spike.

In the comparator circuit, the voltage $V_{THD}$ is responsible for detecting the membrane potential threshold when a spike is coming. The voltage $V_{BIAS}$ controls the bias current in the comparator. When the membrane potential rises above $V_{THD}$, the voltage at the $M8$ gate is decreased and the voltage at the $M5$ gate is increased, generating reset signals. The reset signal is delayed, so the membrane potential $V$ (in the membrane potential circuit at $C_v$) continues to rise beyond $V_{THD}$, up to $V_{dd}$, but as soon as the voltage at $M5$ gate is increased, the membrane potential is reset to $V_c$, which is lower than $V_{THD}$. Next, the voltages at the $M5$ and $M8$ gates return to their resting voltage levels, completing the reset pulses. Transistor $M14$ increases the comparator current during the spike, providing the required amplitude and duration of the reset pulse of the voltage at gate $M8$ \cite{wijekoon2008compact}.

\begin{figure}
    \centering
    \includegraphics[width=0.95\linewidth]{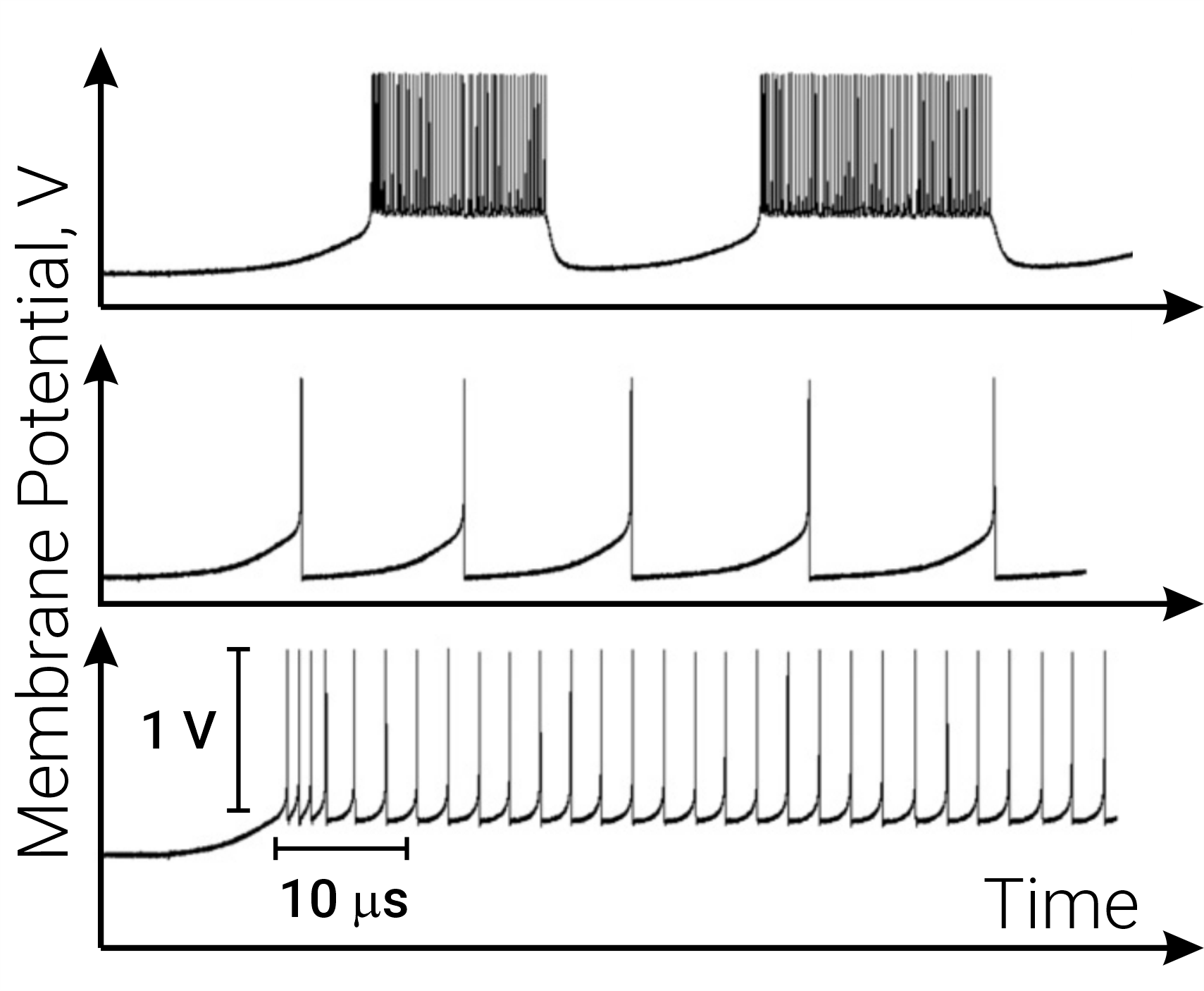}
    \caption{Experimental firing patterns obtained from the fabricated chip containing 202 neurons using 0.35 $\mu m$ CMOS technology. Demonstrated patterns are responses to the input current step for various parameters of bias voltages at node $c$ and node $d$ \cite{wijekoon2008compact}.
    }
    \label{fig:cortical_neuron_pattern}
\end{figure}

The basic circuit comprises 202 neurons with different circuit parameters, including transistor and capacitance sizes. Fabrication was conducted using 0.35 $µm$ CMOS technology. Since the transistors in this circuit operate mainly in strong inversion mode, the excitation patterns are on an "accelerated" time scale, about $10^4$ faster than biological real time. The power consumption of the circuit is less than $10~pJ$ per spike \cite{indiveri2011neuromorphic, wijekoon2008compact}. Main types of characteristic firing patterns of suggested circuit are demonstrated in  figure~\ref{fig:cortical_neuron_pattern}.

\subsection{Transistor-based realisations}

This section is devoted to reviewing such significant projects as SpiNNaker, Neurogrid, TrueNorth,  NorthPole and Loihi. And despite the fact that at the moment of IBM's chip creation there were already such projects as SpiNNaker (2012) \cite{furber2012overview} and Neurogrid (2014) \cite{benjamin2014neurogrid}, we will start with the review of the TrueNorth neuromorphic processor, since in our opinion it was the first hardware implementation of the idea of neuromorphic computing.

\subsubsection{TrueNorth}
The field of brain-inspired technologies was marked in 2015 by the release of the first neuromorphic chip: IBM TrueNorth (figure~\ref{fig:CMOS_chips}a), a neuromorphic CMOS integrated circuit. The TrueNorth chip architecture is based on an organic neurobiological structure, but with the limitations of inorganic silicon technology. The main purpose of this platform was to reproduce the work of existing neural network algorithms of speech and image recognition in real time with minimal energy consumption \cite{akopyan2015truenorth}. This neuromorphic chip contains 4096 neurosynaptic nuclei combined into a two-dimensional array, and contains a total of 1 million neurons and 256 million synapses. The chip has a peak computing performance of $58~GSOPS$ with a power consumption of $400~GSOPS/W$.

The use of the term "neuromorphic" itself implies that the chip is based on an architecture that differs from the familiar von Neumann architecture. Unlike von Neumann machines, the TrueNorth chip does not use sequential programs that map instructions into linear memory. The chip consists of spiking neurons, which are connected in a network and communicate with each other via spikes (voltage pulses). Communication between neurons is tunable, and the data transmitted can be encoded by the frequency, temporal and spatial distribution of the spikes. 
TrueNorth is designed for a specific set of tasks: sensory processing, machine learning and cognitive computing. However, just as in the early days of the computer age and the emergence of the first computer chips, the challenges of creating efficient neurosynaptic systems and optimising them in terms of programming models, algorithms and architectural features are still being solved. Currently, the IBM Truenorth chip is being used by DARPA (Defense Advanced Research Projects Agency) for gesture and speech recognition.

An IBM Research blog post on TrueNorth's performance notes that the classification accuracy demonstrated by the system is approaching the performance of 2016 state-of-the-art implementations, not only for image recognition but also for speech recognition. The \cite{sawada2016truenorth} reports performance data for five digital computing architectures running deep neural networks. A single TrueNorth chip processes 1200-2600 32×32 colour images per second, consuming 170--275~$mW$, yielding an energy efficiency of 6100 -- 7350~$FPS$. TrueNorth multi-chip implementations on a single board process 32×32 colour images at 430 -- 1330 per second and consume 0.89 -- 1.5~$W$, yielding an energy efficiency of $360 -- 1420~FPS/W$. SpiNNaker delivers $167~FPS/W$ while processing 28×28 grayscale images (in a configuration of 48 chips on one board) \cite{stromatias2015scalable}. Tegra K1 GPU, Titan X GPU and Core i7 CPU deliver 45, 14.2 and 3.9~$FPS/W$, respectively, while processing 224×224 colour images \cite{sawada2016truenorth, learning2015performance}.

\subsubsection{SpiNNaker}
The SpiNNaker (Spiking Neural Network Architecture) project, the brainchild of the University of Manchester, was unveiled in January 2012. It is a real-time microprocessor-based system optimised for the simulation of neural networks, and in particular spiking neural networks. Its main purpose is to improve the performance of software simulations \cite{furber2012overview, mayr2019spinnaker}. SpiNNaker uses a custom chip based on ARM cores that integrates 18 microprocessors in 102 $mm^2$ using a 130 $nm$ process. The all-digital architecture uses an asynchronous message-passing network (2-D torus) for inter-chip communication, allowing the whole system to scale almost infinitely. In experiments \cite{painkras2013spinnaker, stromatias2013power, stromatias2015scalable} a 48-chip board (see figure~\ref{fig:CMOS_chips}d) was used, which can simulate hundreds of thousands of neurons and tens of millions of synapses, and consumes about 27-37 $W$ in real time (for different neuron models a network configuration). On average, 2000 spikes formed into a Poisson train were used to encode a digit character from the MNIST dataset, with a classification latency of $20~ms$.

In summary, SpiNNaker is a high-performance, application-specific architecture optimised for tasks from neurobiology and neuroscience in general. It is claimed that the system can also be used for other distributed computing applications such as ray tracing and protein folding \cite{painkras2013spinnaker}. The experimental studies performed suggest that for parallel modelling of deep neural networks, a dedicated multi-core architecture can indeed be energy efficient (compared to competitors and general purpose systems) while maintaining the flexibility of software-implementable models of neurons and synapses.

In 2021 the second generation of neuromorphic chip by the collaboration of Technische University of Dresden and University of Manchester was released -- SpiNNaker2. Its development was conducted within the European Union flagship project ``Human Brain Project''. The overall approach utilised in the creation of SpiNNaker2 has the following parts: keep the processor-based flexibility of SpiNNaker1, don't do everything in software in the processors, use the latest technologies and features for energy efficiency and allow workload adaptivity on all levels \cite{mayr2019spinnaker}. One SpiNNaker2 chip consists of 152 ARM processors (processing elements) arranged in groups of four to quad-processing-elements which are connected by a Network-on-Chip (NoC) to allow scaling towards a large neuromorphic System-on-Chip (SoC). The full-scale SpiNNaker2 will consist of 10 million ARM cores distributed across 70000 сhips in 10 server racks. We'd also highlight that the second generation is designed on a different technical process. Specifically, SpiNNaker1 was realised on $130~nm$ CMOS, while SpiNNaker2 was realised on $22~nm$ FDSOI CMOS \cite{hoppner2021spinnaker}, which allowed not only to increase the performance of a single chip as a whole, but also to improve power efficiency. It is stated that SpiNNaker2 enabling a 10$\times$ increase in neural simulation capacity per watt over SpiNNaker1. Among the potential applications for the SpiNNaker2 the following stand out: naturally, brain research and whole-brain modeling, biological neural simulations with complex plasticity rules (with Spike-timing-dependent plasticity, STDP, for example), low-power inferencing for robotics and embedded AI, large-scale execution of hybrid AI models, autonomous vehicles, and other real-time machine learning applications.

\subsubsection{Neurogrid}
The Neurogrid project, carried out at Stanford University in December 2013 \cite{merolla2013multicast}, is a mixed analogue-digital neuromorphic system based on a 180 nm CMOS process. The project has two main components: software for interactive visualisation and hardware for real-time simulation. The main application of Neurogrid is the real-time simulation of large-scale neural models to realise the function of biological neural systems by emulating their structure \cite{benjamin2014neurogrid}.

The hardware part of Neurogrid contains 16 Neurocores connected in a binary tree: microelectronic chips with a 12x14 $mm^2$ die containing 23 million transistors and 180 pads; a Cypress Semiconductor EZ-USB FX2LP chip that handles USB communication with the host; a Lattice ispMACH CPLD. This makes it possible to establish a link between the data transmitted via USB and the data driven by Neurogrid, and to interleave timestamps with the outgoing data (host binding). A daughter board is responsible for primary axonal branching and is implemented using a field-programmable gate array (FPGA). The Neurogrid board is shown in figure~\ref{fig:CMOS_chips}c. Each Neurocore implements a $256\times256$ silicon neuron array, and also contains a transmitter, a receiver, a router, and two RAMs. A neuron has one soma, one dendrite, four gating variables and four synapse populations for shared synaptic and dendritic circuits \cite{benjamin2014neurogrid}.

The Neurogrid architecture enables to simulate cortical models emulating axonal arbors and dendritic trees: a cortical area is modeled by a group of Neurocores mentioned above and an off-chip random-access memory is programmed to replicate the neocortex’s function-specific intercolumn connectivity \cite{benjamin2021neurogrid}. Therefore,it is indeed simulate the behaviour of large neural structures, including conductance-based synapses, active membrane conductances, multiple dendritic compartments, spike backpropagation, and cortical cell types. No data is available on the power consumption of image recognition, as in the case of TrueNorth or SpiNNaker, but that's understandable: the main purpose of the project is to model the neural activity of living tissues, and that's what it does. One thing we can say for sure is that the neurogrid consumes less power than the GPU for the same simulation -- 120 $pJ$ versus 210 $nJ$ per synaptic activation \cite{knight2021larger}. This confirms the thesis that digital simulation (on CPUs or GPUs), even though it allows solving such problems, but the time and power consumption will be significant compared to specialised architectures.

\subsubsection{NorthPole}
\begin{figure}
    \centering
    \includegraphics[width=0.95\linewidth]{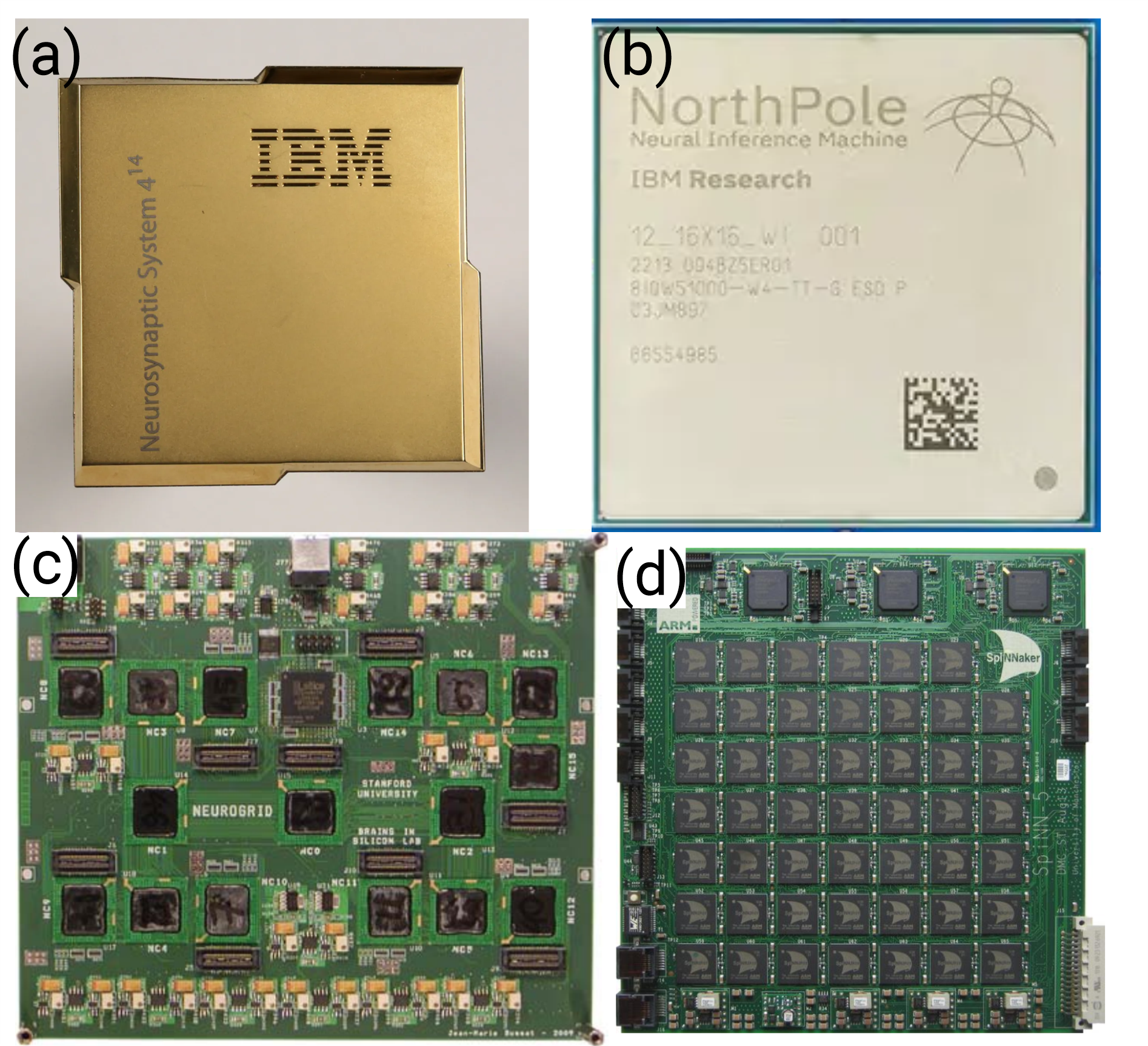}
    \caption{
    Images of some CMOS spiking neural network realisations: (a) IBM TrueNorth chip \cite{akopyan2015truenorth}, (b) IBM NorthPole chip \cite{modha2023neural}, (c) Neurogrid board \cite{merolla2013multicast}, (d) SpiNNaker (spiking neural network architecture), a massively parallel, manycore supercomputer architecture \cite{stromatias2013power}.
    }
    \label{fig:CMOS_chips}
\end{figure}

A striking example of a hardware implementation of a semiconductor bio-inspired neural network is the recently (2023) introduced NorthPole chip from IBM \cite{modha2023neural} (figure~\ref{fig:CMOS_chips}b). NorthPole is an extension of TrueNorth, and it's not surprising that it inherits some of the technology used there. The NorthPole architecture is designed for low-precision, common-sense computing while achieving state-of-the-art inference accuracy for neural networks. It is optimised for 8-, 4- and 2-bit precision, eliminating the need for high precision during training. The NorthPole system consists of a distributed modular array of cores, with each core capable of massive parallelism, performing 8192 2-bit operations per cycle. Memory is distributed between and within the cores, placing it in close proximity to the computation. This proximity allows each core to take advantage of data locality, resulting in improved energy efficiency. NorthPole also incorporates a large on-chip memory area that is neither centralised nor hierarchically organised, further enhancing its efficiency. 

Potentially, the NorthPole chip opens up new ways for the development of intelligent data processing for tasks such as optimisation (of systems, algorithms, scalability, etc.), for image processing for digital machine vision, and data recognition for autopilots, medical applications, etc. Also the chip was run in such well-known tests as ResNet-50 image recognition and YOLOv4 object detection models, where it showed outstanding results: higher energy and space efficiency, and lower latency than any other chip available on the market today, and is roughly 4000 times faster than TrueNorth (since the requirements to the accuracy of calculations in the chip are reduced, it is not possible to correctly estimate the specific performance). The article \cite{modha2023neural} provides the following data: NorthPole based on 12~$nm$ node processing technology delivers 5 times more frames per joule than GPU NVIDIA H100 based on 4~$nm$ technological process (571 vs. 116 $F/J$) and 1.5 times more than the specialised for neural network use Qualcomm Cloud AI 100 based on 7~$nm$ technological process. The reason for this is the locality of the computation -- by eliminating off-chip memory and intertwining on-chip memory with compute memory, the locality of spatial computation is ensured and, as a result, energy efficiency is increased. Also low-precision operations further increase NorthPole's lead over its competitors. 

In original paper \cite{modha2023neural} D.S. Modha and his team tested NorthPole only for use in computer vision. However, with this sort of potential, this chip can also be used for image segmentation and video classification. According to the information on the IBM's blog it was also tested in other areas, such as natural language processing (on the encoder-only BERT model) and speech recognition (on the DeepSpeech2 model).

\subsubsection{Loihi and Loihi2}
\begin{figure}
    \centering
    \includegraphics[width=0.9\linewidth]{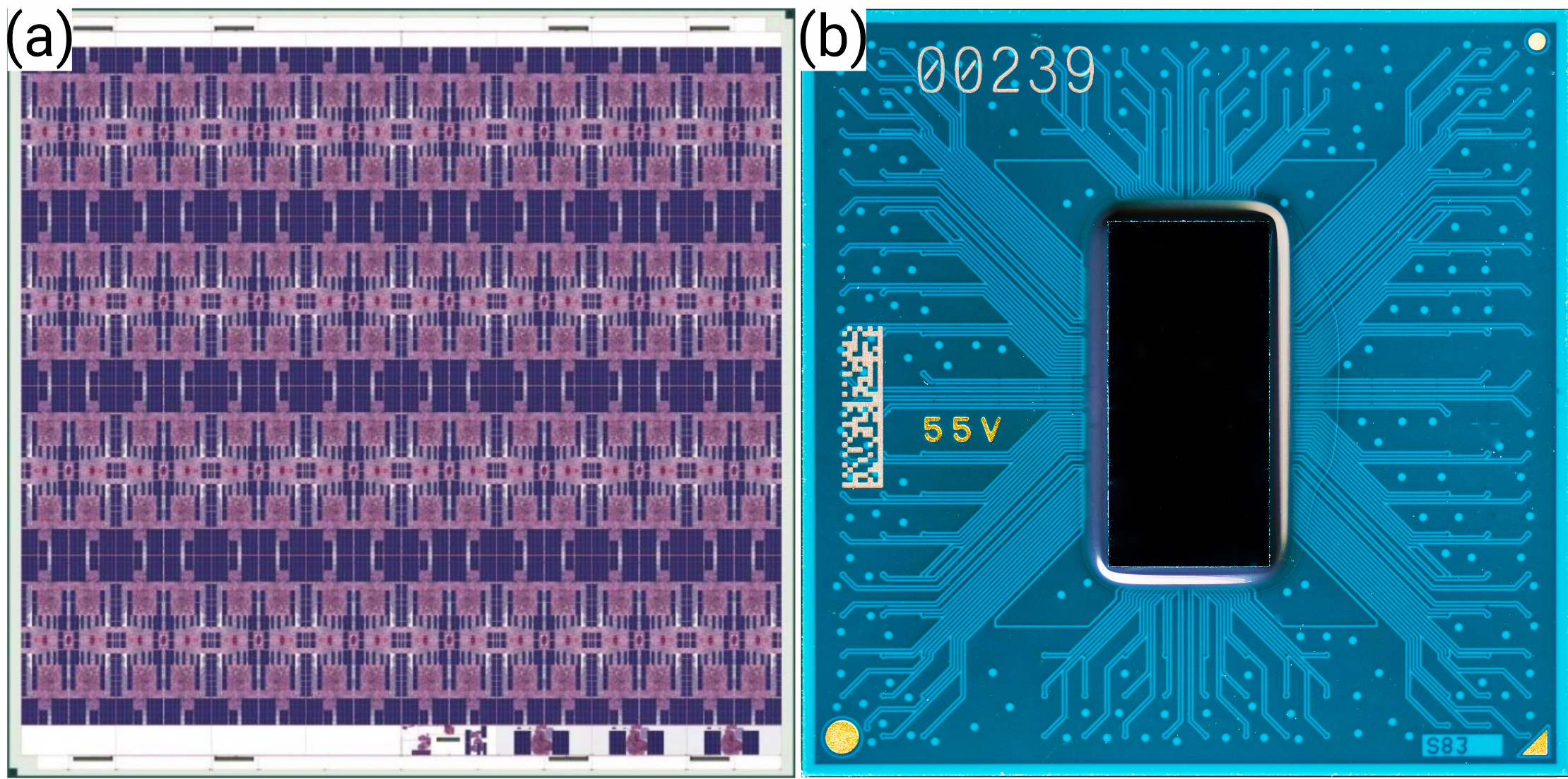}
    \caption{(a) Loihi chip plot and (b) Intel's the second-generation neuromorphic research chip Loihi 2 (pictures from the Intel's official website from the News about "Intel Advances Neuromorphic with Loihi 2, New Lava Software Framework and New Partners")
    }
    \label{fig:Loihi}
\end{figure}

In 2018, Intel Labs unveiled the first neuromorphic manycore processor that enables on-chip learning and aims to model spiking neural networks in silicon. The name of this processor is Loihi (figure~\ref{fig:Loihi}a). Technologically, Loihi is a 60~$mm^2$ chip manufactured in Intel's FinFET 14~$nm$ process. The chip instantiates a total of 2.07 billion transistors and 33 MB of SRAM across its 128 neuromorphic cores and three x86 cores to manage the neuro cores and control spike traffic in and out of the chip. It supports asynchronous spiking neural network models for up to 130000 synthetic compartmental neurons and 130 million synapses. Loihi's architecture is designed to enable the mapping of deep convolutional networks optimised for vision and audio recognition tasks. Loihi was the first of its kind to feature on-chip learning via a microcode-based learning rule engine within each neural core, with fully programmable learning rules based on spike timing. Intel's chip allows the SNN to incorporate: 1) stochastic noise, 2) configurable and adaptable synaptic, axonal and refractory delays, 3) configurable dendritic tree processing, 4) neuron threshold adaptation to support intrinsic excitability homeostasis, and 5) scaling and saturation of synaptic weights to support ``permanence'' levels beyond the range of weights used during inference \cite{davies2018loihi, lin2018programming}. 

The on-chip learning is organised in such a way that the minimum of the loss function over a set of training samples is achieved during the training process. Also, learning rules satisfies the locality constraint: each weight can only be accessed and modified by the destination neuron, and the rule can only use locally available information, such as the spike trains from the pre-synaptic (source) and post-synaptic (destination) neurons. Loihi was the first fully integrated digital SNN chip that supported diverse local information for programmable synaptic learning process such as: 1) spike traces corresponding to filtered presynaptic and postsynaptic spike trains with configurable time constants, 2) multiple spike traces for a given spike train filtered with different time constants, 3) two additional state variables per synapse, besides the normal weight, to provide more flexibility for learning and 4) reward traces that correspond to special reward spikes carrying signed impulse values to represent reward or punishment signals for reinforcement learning \cite{davies2018loihi}. 

In Intel Labs original work the performance of Loihi was checked on the LASSO task ($l_1$-minimising sparse coding problem). The goal of this task is to determine the sparse set of coefficients that best represents a given input signal as a linear combination of features from a feature dictionary \cite{davies2018loihi}. This task was solved on a $52\times52$ image with a dictionary of 224 atoms and during that the Loihi allowed to provide 18 times compression of synaptic resources for this network. The sparse coding problem was solved to within 1\% of the optimal solution. Unfortunately, the article does not provide data on energy efficiency and speed of the chip when solving this problem. A bit later in \cite{lin2018programming} the classical classification task on MNIST dataset was solved with accuracy of 96.4\%.

On 30 September in 2021 Intel Labs introduced their second-generation neuromorphic research chip Loihi 2 (figure~\ref{fig:Loihi}(b)), as well as the open-source software framework LAVA for developing neuro-inspired applications. 
Based on the Loihi experience, Loihi 2 supports new classes of neuro-inspired algorithms and applications, while providing up to 10 times faster processing, up to 15 times greater resource density with up to 1 million neurons per chip, and improved energy efficiency \cite{davies2021taking}. Loihi 2 has the same base architecture as its predecessor Loihi, but with new functionality and improvements. For example, the new chip supports fully programmable neuron models with graded spikes (Loihi supported only generalised LIF neuron model with binary spikes, whereas Loihi 2 supports any you like models) and achieves a 2$\times$ higher synaptic density. The \cite{shrestha2024efficient} reports a comparison of the performance and energy efficiency of the Loihi 2 chip compared to the performance of the the NVIDIA Jetson Orin Nano on video and audio processing task. Computing neuromorphic systems based on the Loihi 2 provide significant gains in energy efficiency, latency, and even throughput for intelligent signal processing applications (such as navigation and autopilot systems, voice recognition systems) compared to conventional architectures. For example, the Loihi 2 implementation of the Intel NSSDNet (Nonlocal Spectral Similarity-induced Decomposition Network) increases its power advantage to 74$\times$ compared to NsNet2 (Noise Suppression Net 2) running on the Jetson Orin Nano platform. The Loihi 2 has also demonstrated their advantages in Locally Competitive Algorithm implementation \cite{parpart2023implementing}. Loihi 2 is also capable of reproducing bio-realistic neural network implementation and it is flexible in terms of supporting different neuron models. In \cite{uludaug2023bio} authors demonstrate a showcase of implementing a simplified bio-realistic basal ganglia neural network that carries ``Go/No-Go'' task, by using Izhikevich neurons. 

\subsection{Memristor-based realisations}

A memristor is the fourth passive element of electrical circuits.
This element is a two-terminal device, the main property of which is the ability to memorise its state depending on the applied bias current. Theoretically proposed by Leon Chua in 1971 \cite{Chua1971Memristor}, the memristor has seen the daylight as a prototype (memory + resistor) based on a thin film of titanium dioxide in 2008 thanks to HP Labs company \cite{Strukov2008memristor}. The reason for the development of this device in the first place was the following problem: to further improve the efficiency of computing, electronic devices must be scalable to reduce manufacturing costs, increase speed and reduce power consumption -- that is, more and more transistors must be placed on the same area of the crystal each time. However, due to physical limitations and rising manufacturing costs when moving to new process standards (to a 10 nm process, for example), the processing nodes of a traditional CMOS transistor can no longer scale cost-effectively and sustainably. As a result, new electronic devices with higher performance and energy efficiency have become necessary to satisfy the needs of the ever-growing information technology market \cite{li2018review}, implementing new ``non-von Neumann'' paradigms of in-memory computation \cite{Erokhin2020Memristive,Lee2020Nanoscale}. 

\begin{figure}
    \centering
    \includegraphics[width=0.75\linewidth]{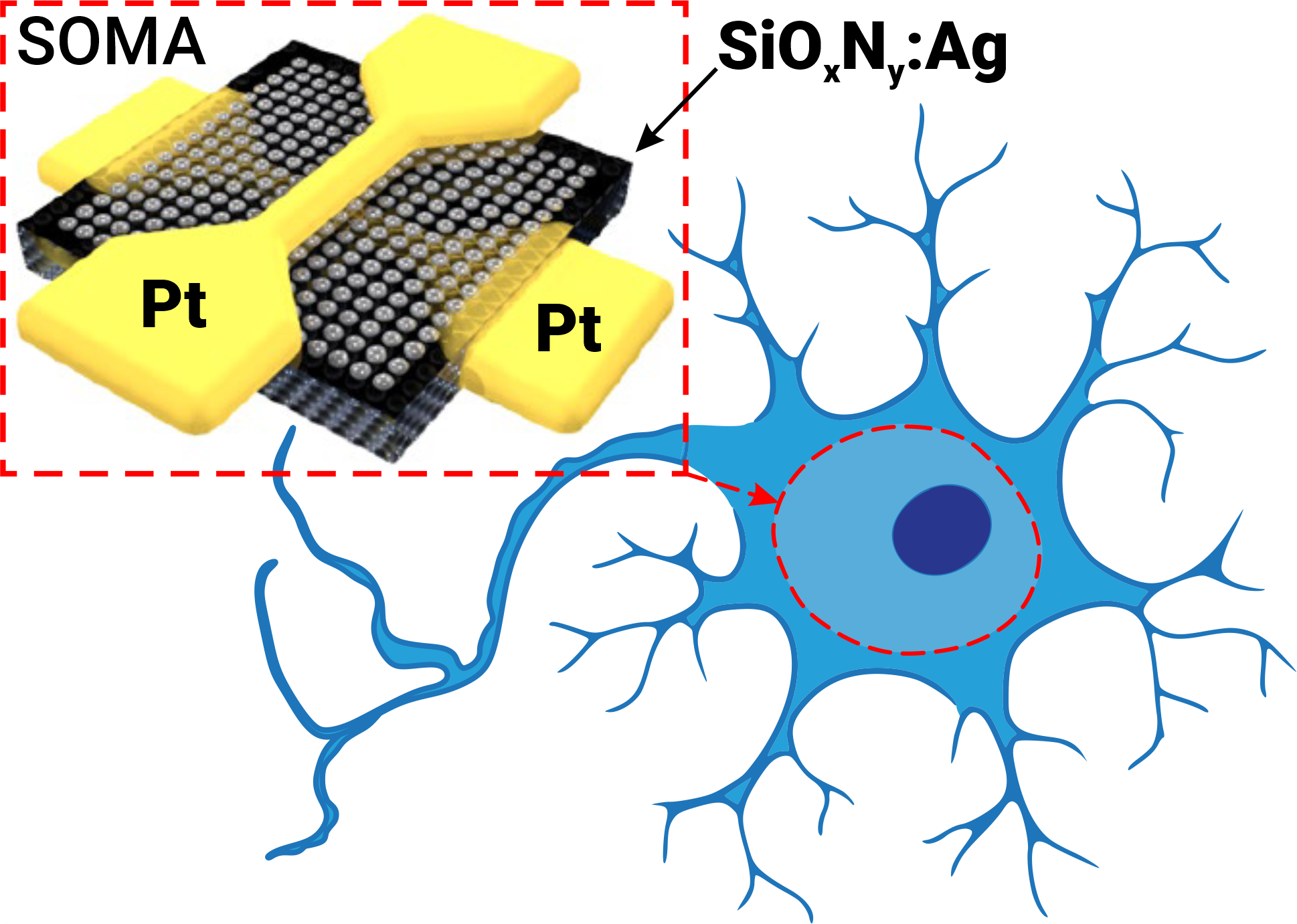}
    \caption{Illustration of the artificial neuron based on a diffusive memristor, consisting of a $SiO_x N_y:Ag$ layer between two Pt electrodes \cite{wang2018fully}. } 
    \label{fig:memr_neuron}
\end{figure}

Initially, memristive systems acted as elements of energy-efficient Resistive Random-Access Memory memory (RRAM) by using two metastable states with high and low resistance, switching between which is carried out by applying an external voltage. However, in recent years, the potential application of memristors can be used to realise the functions of both synapses and neurons in both ANNs and SNNs \cite{Mikhaylov2023Supercomputing}. Figure~\ref{fig:memr_neuron} shows one of the realisations of a memristor-based neuron. It exploits the diffusion processes between two types of layers: dielectric $SiO_x N_y:Ag$ layer (doped with $Ag$ nanoclusters) and $Pt$ layer \cite{wang2018fully}. The $SiO_x N_y:Ag$ material serves as the functional part of the memristor, allowing the creation of a model of the leaky integrate-and-fire neuron. The diffusive memristor integrates the presynaptic signals (arriving at one of the $Pt$ electrodes) within a time window and transitions to a low resistance state only when a threshold is reached.

Using nanotechnology capabilities, it is possible to miniaturise memristive electronic devices to units of nanometers, which allows achieving a high density of elements on a chip. An important advantage of memristors is their compatibility with CMOS technology: memristors can be organised fairly easily into a crossbar architecture that takes advantage of parallel in-memory computing. With nanoscale, two-terminal and semiconductor memristor, memristor crossbars are characterised by high element density and better energy efficiency than their transistor counterparts. Due to this, this approach makes it possible to implement memristive networks based on 2D- and 3D-integrated crossbar structures to increase the speed of signal transmission \cite{Strukov2009dimensional, Jo2010Nanoscale, likharev2011crossnets, prezioso2015training, prezioso2016self, Papandroulidakis2017Crossbar}. This is why memristor crossbars are seen as a good candidate as a basis for neuromorphic networks \cite{zhang2018neuromorphic}.

It is assumed that due to similar mechanisms of ion exchange in biological nerve cells and memristors, the last ones can mimic synapses and even neurons with sufficiently high accuracy. Since the capabilities of a neural network are usually determined by its size (number of neurons and inter-neuron connections), a scalable and energy-efficient component base is required to create a more powerful and energy-efficient system \cite{xia2019memristive,Tzouvadaki2023Interfacing}. There have been many works aimed at studying the properties of memristors in terms of their application as synaptic elements and neurons, and there have even been successes in demonstrating the effect of synaptic plasticity \cite{la2015filamentary} and the operation of the integrate-and-fire neural model \cite{zhang2017artificial}. Also an artificial neuron based on the threshold switching and fabricated on the basis of $Nb O_x$ material, has been demonstrated. Such a neuron displays four critical features: threshold-driven spiking, spatiotemporal integration, dynamic logic and gain modulation \cite{duan2020spiking}.

The dynamic characteristics of memristors in combination with their nonlinear resistance allow us to observe the responses of the system to external stimulation \cite{Pershin2019Dynamical}. At the same time, the stochastic properties of memristors due to interaction with the external environment \cite{Carboni2019Stochastic} can be used both for controlling metastable states and for hardware design of SNN circuits \cite{Makarov2022Toward}. The first experimental prototypes of SNNs based on memristors and CMOS have already been created (synapses are realised via memristor crossbars, and neurons - via semiconductor transistors) and they can ``to a certain extent'' emulate spike-timing dependent plasticity \cite{Prezioso2018Spike,Demin2021Necessary}. However, they are based on a simplified concept of synaptic plasticity based on overlapping pre- and postsynaptic adhesions \cite{Demin2021Necessary}, which has led to reduced energy efficiency and rather complex technological design of SNN design circuits. Currently, approaches to overcome these problems are being developed, for example, on the basis of complete rejection of analogue-digital and digital-analogue transformations and creation of neuromorphic systems in which all signal processing occurs in analogue form \cite{Amirsoleimani2020Inmemory} or by creating the concept of self-learning memristor SNNs.


Unfortunately, to realise a fully-memristive neuromorphic neural network it is necessary to achieve a non-linear dynamic signal processing. This challenge is so far not overcome by memristors alone, and every memristive neural network proposal relies on the operation of classical semiconductor transistors in one way or another. Below we consider two interesting realisations of CMOS memristive neural networks.

\subsubsection{Neuromorphic network based on diffusive memristors}

In 2018 Zhongrui Wang and his colleagues presented an artificial neural network implemented on diffusive memristors (and also on transistors) that is capable to solve pattern classification task with unsupervised learning \cite{wang2018fully}. Note that the network demonstrated in the paper, recognised only four letters ("U", "M", "A", "S"), presented as 4-by-4 pixel images. Nevertheless, a working memristive network is a significant achievement and shows the feasibility in principle of the idea of memristor-based neuromorphic hardware computing. In  figure \ref{fig:memr_neuron} demonstrated the central part of this network and artificial neuron in particular -- diffusive memristor consisted of a $Si O_x N_y: Ag$ layer between two $Pt$ electrodes and serving as the neuron's soma.

The most important difference between a diffusion memristor and a traditional memristor is that once the voltage is removed from the device terminals, it automatically returns to its original high-resistance state. The dynamics of the diffusion process in a diffusion memristor has similar physical behaviour to biological $Ca^{2+}$ dynamics, which can accurately mimic different temporal synaptic and neuronal properties \cite{xia2019memristive}.

\begin{figure}
    \centering
    \includegraphics[width=0.9\linewidth]{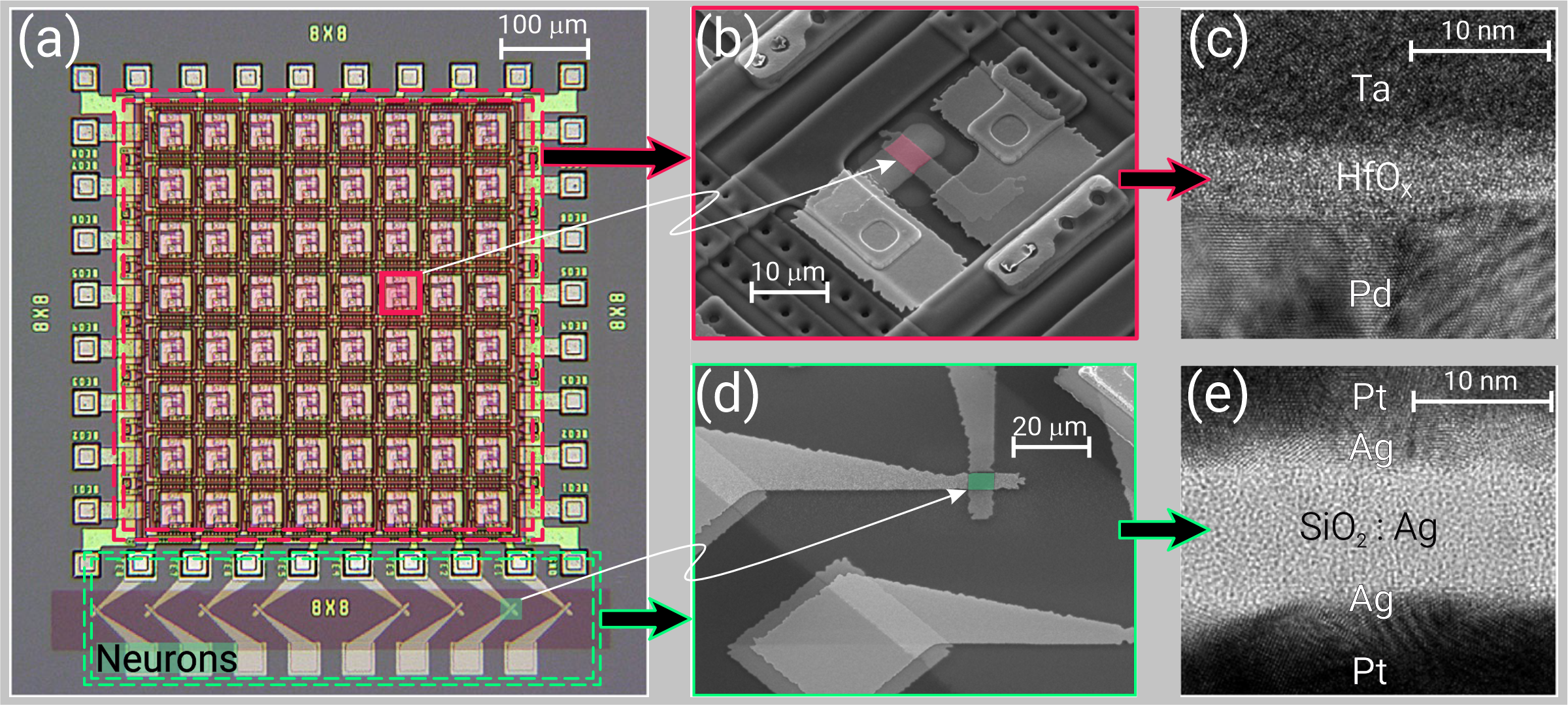}
    \caption{(a) Microphotograph of the fully memristive spiking neural network, consisting of memristive synapse crossbar (in red frame) and memristive neurons (in blue frame). (b) SEM microphotograph of a single memristive synapse. (c) TEM (transmission electron microscopy) image of the synapse cross-section with the structure of drift memristor, $Pd / HfO_x / Ta$. (d) SEM microphotograph of a single memristive neuron. (e) TEM image of the neuron cross-section with the structure of diffusive memristors, $Pt/Ag/SiO_x: Ag/Ag/Pt$ \cite{wang2018fully}.
    }
    \label{fig:memr_chip}
\end{figure}

The diffusive memristor in the artificial neuron is very different from non-volatile drift memristors or phase-change memory devices used as long-term resistive memory elements or synapses. 
The point is that the memristor processes the incoming signals within a certain time window (characteristic time, which in the diffusive memristor model is determined by the Ag diffusion dynamics to dissolve the nanoparticle bridge and return the neuron to its resting state) and then, only when a threshold has been reached, transitions to a low-resistance state. 
Figure~\ref{fig:memr_chip}(a) shows an integrated chip of the memristive neural network \cite{wang2018fully}, consisting of a one-transistor–one-memristor (figure~\ref{fig:memr_chip}(b,c)) synaptic 8-by-8 array and eight diffusive memristor neurons (figure~\ref{fig:memr_chip}(d,e)). The synapses were created by combining drift memristors with arrays of transistors. In this configuration, each memristor (i.e., the $Pd/HfO_2/Ta$ structure) is connected to a series of n-type enhancement-mode transistors. When all the transistors are in an active state, the array functions as a fully connected memristor crossbar.

The input images are divided into four 2-by-2 pixel subsets, where each pixel is assigned a specific pair of voltages (equal in modulus but different in sign), depending on the colour and intensity of the pixel. 
Each resulting subset is expanded into a single-column input vector comprising eight voltages. This input vector is then applied to the network, which consists of eight rows, at each time step. For each possible subset, there is a corresponding convolution filter implemented using eight memristor synapses per column. In total, there are eight filters in an 8 × 8 array. 
The convolution of the eight filters for each sub-image simultaneously results in the "firing" of the corresponding neurons, which fulfil the role of ReLU. 
Eight outputs of these "ReLUs" coupled with the 8-by-3 fully connected memristive crossbar, whose fan-out concentrates all spikes on the last layer of the neural network consisting of 3 neurons ''firing'' which corresponds to a certain recognised image. It is claimed that the neuromorphic nature of the presented network is ensured due to the realisation of the effect of spike-timing-dependent plasticity (STDP) used in the learning process and reflected in the change in the resistance of memristors in the crossbar when the next voltage pulse is passed. Despite the insufficient (in our opinion) representativeness of the learning results, the network architecture and the physics behind the used devices do not allow us to doubt the presence of the STDP effect, considering that similar effects were observed in memristors back in 2015 \cite{saighi2015plasticity}.

\subsubsection{STDP learning in partially memristive SNN}

\begin{figure}
    \centering
    \includegraphics[width=0.9\linewidth]{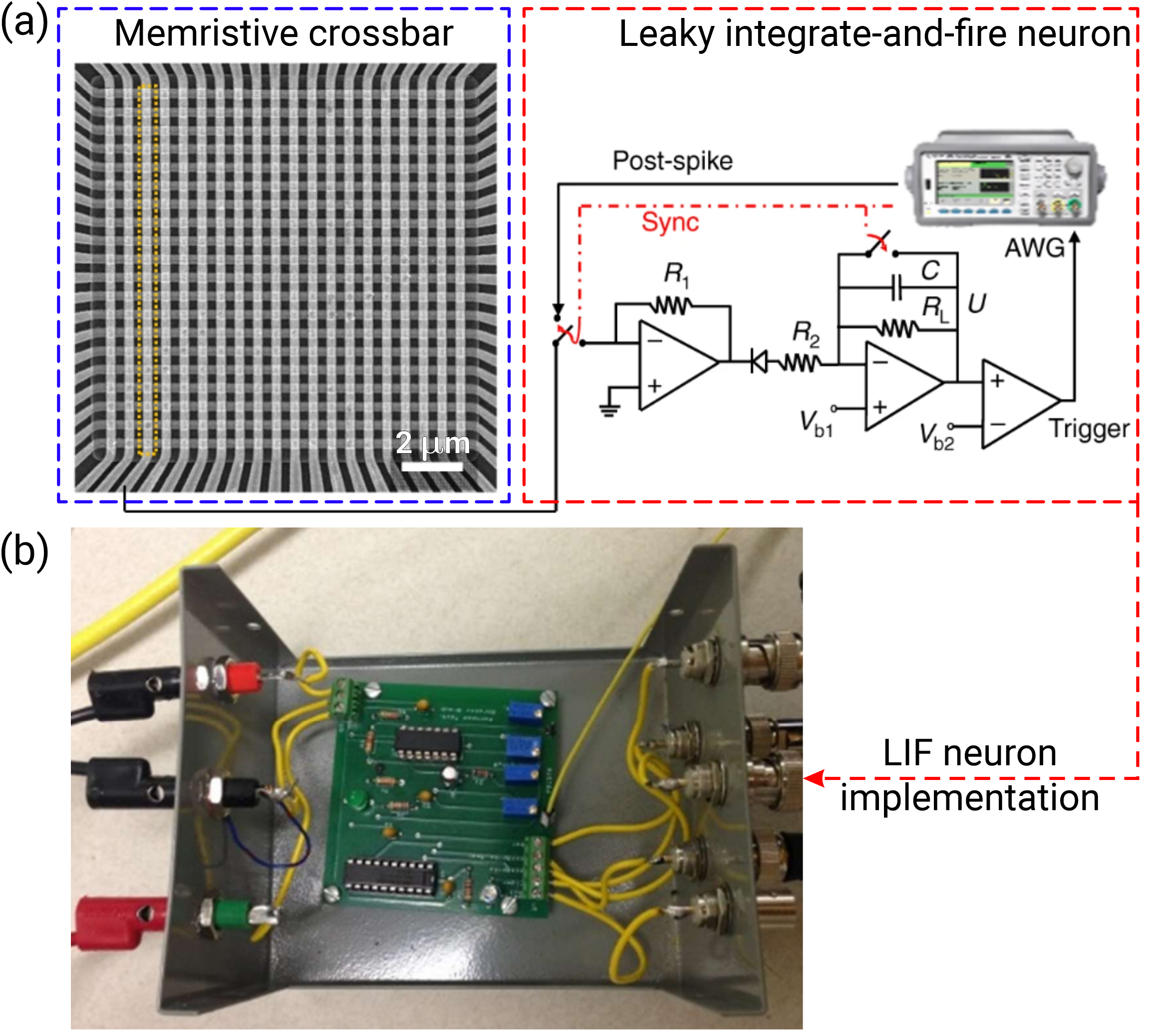}
    \caption{a -- The SEM top view image of the 20 memristive crossbar-integrated synapses connected to a single leaky-integrate-and-fire (LIF) neuron (schematic is shown). b -- The experimental setup of the LIF neuron implementation \cite{Prezioso2018Spike}.
    }
    \label{fig:Prezioso}
\end{figure}

The next realisation of the memristive-based neural network is proposed by M. Prezioso et al., who experimentally demonstrated operation and STDP learning in SNN, implemented with the passively integrated (0T1R) memristive synapses connected to a silicon leaky integrate-and-fire (LIF) neuron \cite{Prezioso2018Spike}. The experimental setup consisted of 20 input neurons connected via 20 memristive crossbar-integrated synapses to a single LIF neuron. The input neurons -- one neuron for each row of the memristive crossbar -- are implemented using the off-the-shelf digital-to-analogue converter circuits. Synapses are implemented by $Pt/Al_2 O_3 / Ti O_{2-x}/Pt$ based memristors in a $20\times20$ crossbar array. In this SNN implementation there is only a single LIF output neuron that connected to the third array column, while the other columns are grounded. LIF neuron is also realised on the custom-printed circuit board (see figure~\ref{fig:Prezioso}). This arrangement allows connecting the crossbar lines either to the input/output neurons during network operation or to a switch matrix, which in turn is connected to the parameter analyser, for device forming, testing, and conductance tuning. Once the threshold is reached, the LIF neuron fires an arbitrary waveform generator. The characteristic duration of voltage spikes is on the order of 5 ms.

The neural network was trained to solve the coincidence detection problem (the task of identifying correlated spiking activity) using the STDP learning mechanism, which allows training a single neuron to produce an output pulse when it receives simultaneous (i.e., correlated) pulses from multiple input neurons. The relevance of such a task is due to the fact that the coincidence detection mechanism is an integral part of various parts of the nervous system, such as the auditory and visual cortex, and is generally believed to play a very important role in brain function \cite{softky1993highly, mainen1995reliability, stevens1998input}.

The conducted experiment has confirmed one of the main challenges for SNN implementation with memristors -- their device-to-device variations. This imposes its own limitations on STDP learning, where it is important to consider certain time intervals (windows). Ideally, all synapses should be identical, and conductance updates for them should occur in equal STDP windows. In reality, however, conductance updates differ significantly between different memristive synapses, precisely because of differences in device switching thresholds. SNNs usually operate with spikes of the same magnitude, which allows to realise parallel updating of weights in several devices in crossbar circuit designs. In this experiment, the magnitude of the spikes was chosen based on the average switching threshold of the devices. Therefore, the change of conductivity in devices with a larger switching threshold was naturally smaller. 

So, CMOS models provides great integration capabilities, while at the same time requiring high power supply and a large number of auxiliary elements to achieve bio-realism. Also for this technology it is still difficult to achieve high parallelism in the system due to significant interconnect losses. Meanwhile, important applications based on semiconductor neuromorphic chips are already being solved, and within the framework of the ``Human Brain Project'', the second generation of the spiking neural network architecture SpiNNaker has even been created. 
Creating and advancing higher performance and energy efficient semiconductor chips is a very important issue. There are broad areas where these devices may be in high demand, primarily because of the possibility of mobile deployment from smart bioprostheses and brain-computer interfaces to ``thinking'' robot-androids.

Memristors can achieve bio-realism with fewer components (the ionic nature of their functioning is similar to that in neurons), but the technology of their manufacturing has not yet been refined and the problem of device-to-device variations has not been solved. Therefore, at the moment, we should assume that memristors will be an auxiliary tool for semiconductor technology. It is not realistic to create a fully memristive neuromorphic processor separately, but maybe it is not necessary?

At the device level, the energy required for computation and weight updates is minimal, as everything is rooted in the presence or absence of voltage spike. At the architecture level, computation is performed directly at the point of information storage (neuromorphism and rejection of von Neumann architecture), avoiding data movement as is the case in traditional digital computers. In addition, potentially memristor networks have the ability to directly process analogue information from various external sensors and sensing devices, which will further reduce processing time and power consumption. The superconductor element base used to build neuromorphic systems, discussed below, has the same capability. Experimental implementation of large memristor neural networks used for working with real data sets is at an early stage of development compared to CMOS analogues \cite{xia2019memristive}. And the device-to-device variation in memristors’ switching thresholds is still the major challenge.


\section{Superconductor-based bio-inspired elements of neural networks}

Superconducting digital circuits are also an attractive candidate for creating large-scale neuromorphic computing systems \cite{ishida2021superconductor, semenov2021new, semenov2023biosfq}. Their niche is in tasks that require both high performance and energy-efficient computing. Modern superconducting technologies make it possible to perform logical operations at high frequencies, up to 50~$GHz$, with energy consumption in the order of $10^{-19} - 10^{-20} J$ per operation \cite{herr2011ultra, mukhanov2011energy, tanaka2012, soloviev2017beyond, semenov2017ac, ishida2021superconductor}. Superconducting devices \cite{Mukh2} lend themselves to the creation of highly distributed networks that offer greater parallelism than the conventional approaches mentioned above. For example, in a crossbar-based synaptic network, the resistive interconnection leads to performance degradation and self-heating. In contrast, the zero-dissipation superconducting interconnection at cryogenic temperatures provides a way to limit interconnection losses. 

The main problem with superconductor-based technologies is their relatively low scalability. Recently, the size of superconducting logic elements, the main part of which is the Josephson junction (JJ) \cite{Josephson_1962, Anderson_1963}, is about 0.2~$\mu m^2$ (approximately $10^7$ JJs per square centimeter) \cite{tolpygo2018super, tolpygo2021inductance, tolpygo2024development}. This value is comparable to the size of a transistor at 28~$nm$ process technology in 2020. At the same time, superconducting hardware has a competitive advantage over CMOS technology due to its ability to exploit the third dimension in chip manufacturing processes. Incorporating vertically stacked Josephson inductors fabricated with self-shunted Josephson junctions in SFQ-based circuits would increase circuit density with minimal impact on circuit margins \cite{castellanos2019stacked}. In addition, the operating principles of Josephson digital circuits, which manipulate magnetic flux quanta with associated picosecond voltage pulses, are very close to the ideas of spiking neural networks.


Further development of the superconducting implementation of neuromorphic systems thus offers the prospect of creating a neuronet that emulates the functioning of the brain with ultra-high performance, and at the same time the neurons themselves prove to be relatively compact, since it is enough to use only a few heterostructures for the proper functioning of the cell.

The design of neurons and synapses using Josephson junctions (JJs) and superconducting nanowires is discussed in the following subsections.

\subsection{Implementations based on Josephson junctions}

This approach is based on the quantisation of magnetic flux in superconducting circuits. Moreover, the flux quantum \cite{Roditchev_2015, Stolyarov_2018, Grbenchuk_2020} can only enter and exit through a weak link in the superconductor: the Josephson junction, an analogue of an ion-permeable pore in the membrane of a biological neuron. This fact is important because studies of neural systems focus mainly on studying and reproducing the behaviour of neurons, including complex patterns of neuronal activity, their ''firings''.
Just as biological neurons have a threshold voltage above which an action potential is generated, a Josephson junction has a threshold current, the Josephson junction critical current, $I_c$. When the current through the junction exceeds this value, a voltage spike-like pulse is generated. 

One of the most common ways to describe Josephson junctions is the resistively shunted junction model, 
where the junction can be replaced with an equivalent circuit of three parallel elements: a Josephson junction, a resistor and a capacitor \cite{soloviev2017beyond}. 
The current through the Josephson junction can be written as the sum of three currents:
\begin{equation}
     I_{S} + I_{N} + I_{D}  = I,
\end{equation}
where $I_{S}$ is the supercurrent governed by the Josephson phase $\varphi$ of the JJ (the current phase relation, CPR, is an important property of the heterostructure), $I_{N}$ is the normal component of the current, for which Ohm's law can be applied, and $I_{D}$ is the capacitive component of the current. Based on the Josephson relations for current components, we obtain a 2nd order differential equation describing the phase dynamics of the Josephson junction:

\begin{equation}
     I_{c} \sin\varphi+\frac{\hbar}{2eR} \dot{\varphi}+\frac{\hbar C}{2e}\ddot{\varphi} = I,
\end{equation}
where $C$ and $R$ are the capacitance and the normal resistance of the Josephson junction.

This equation is identical to that of a forced damped pendulum, where the first term gives the torque due to a “gravitational” potential, the second term is the damping term and the third term is the kinetic energy with a mass promotional to $C$. The sum of all three terms is the external torque corresponding to the applied current. When a small torque below the critical value is applied, the phase $\varphi$ will increase and reach a static value. As the phase gets close to $\pi$, there will be a probability that the pendulum will go over the potential energy maximum, resulting in the emission of a single flux quantum (SFQ) pulse. When the applied current exceeds the critical current, the torque is enough to continually drive the pendulum over the potential energy maximum generating a series of SFQ pulses. 

\begin{figure}
    \centering
    \includegraphics[width=1\linewidth]{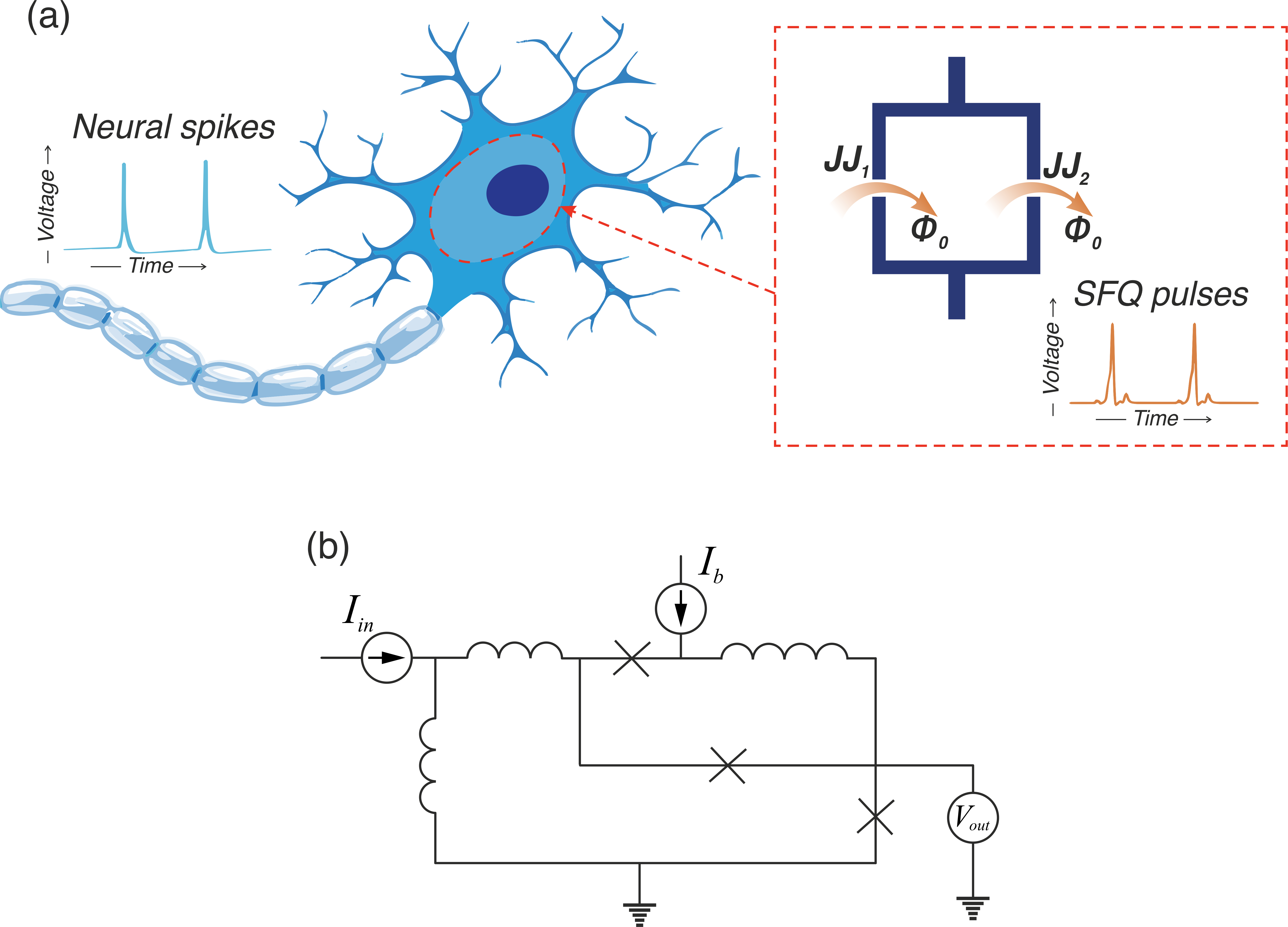}
    \caption{
(a) A sketch explaining the concept of the Josephson neuron: a superconducting circuit plays the role of a membrane impermeable to magnetic flux quanta; weak points of Josephson junctions allow quanta to enter (and exit) the circuit. (b) By replacing one of the junctions with an asymmetric quantum interferometer (SQUID) one enables control of the ratio between the ``widths'' of the input and output channels.}
    \label{fig:3JJ}
\end{figure}

The Josephson junction soma (figure~\ref{fig:3JJ}a), or JJ soma, proposed by Crotty et al. \cite{crotty2010josephson}, is a circuit of two Josephson junctions connected in a loop which displays very similar dynamics to the Hodgkin-Huxley model. The two junctions behave phenomenologically like the sodium and potassium channels, one allowing magnetic flux to charge up the loop and the other allowing flux to discharge from the loop. The circuit exhibits many features of biologically realistic neurons, including the evocation of action potentials ``firing'' in response to input stimuli, input strength thresholds below which no action potential is evoked, and refractory periods after ``firing'' during which it is difficult to initiate another action potential.

Articles (\cite{skryabina2022superconducting, schegolev2023bio, karimov2024magnetic}) discuss an improved version of such a soma, where the input ``pore'' is represented as an asymmetric interferometer (figure~\ref{fig:3JJ}b). The main advantage of the 3JJ neuron is that its mode of operation is easier to control. It has been shown that the 3JJ neuron has a much wider range of parameters in which switching between all operating modes (bursting, regular, dead, injury) is possible simply by controlling the bias current. Furthermore, the 3JJ neuron can be made controllable using identical Josephson junctions, and this design tolerates larger variations in the physical parameters of the circuit elements.

\begin{figure}
    \centering
    \includegraphics[width=1.0\linewidth]{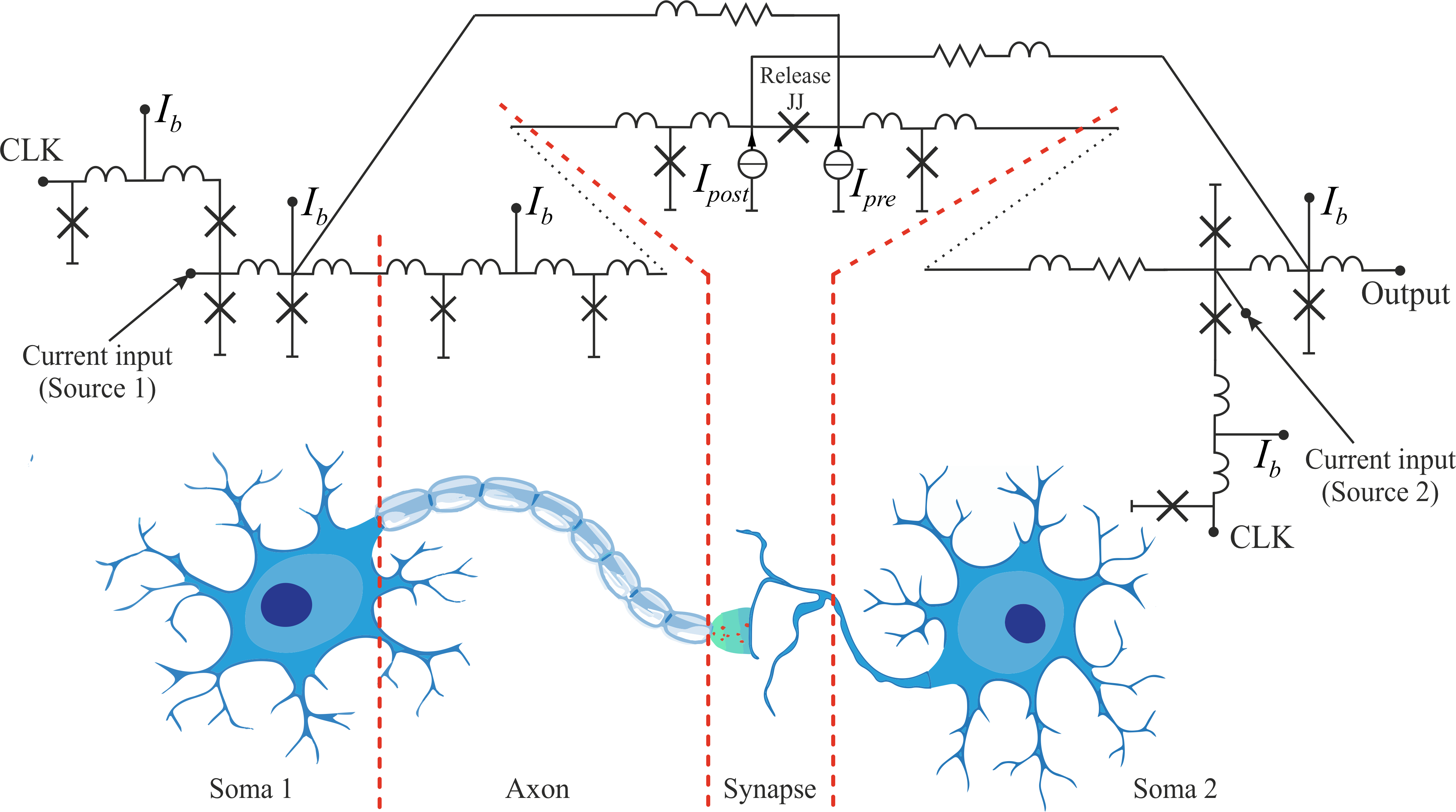}
    \caption{Schematic representation of the complete RSFQ-based architecture, along with a comparison of each of its components to the biological prototype. Here the JTLs act as axons, the Josephson junction in the centre as soma. }
    \label{fig:complete_rsfq}
\end{figure}

Each element of the nervous system can be represented by a similar behavior element of classical RSFQ logic \cite{feldhoff2021niobium}. Complete architecture and comparison with the biological archetype are shown in figure~\ref{fig:complete_rsfq}. This circuit is driven by clocking regime in order to maintain stability when unforeseen side effects occur. One of the most compelling capabilities of superconducting electronics is its ability to support very high clock rates. 

\begin{enumerate}
    \item Soma -- Josephson Comparator (JC).    
    This is a clocked decision element that decides to let a single flux quantum pass in response to a current driven into source 1 in figure~\ref{fig:complete_rsfq}. The decision function of the comparator is similar to the activation of a neuron in response to a current stimulus. Note that the Josephson comparator has been widely used as a nonlinear element in superconducting spiking neural networks \cite{harada1991artificial, yamanashi2012pseudo}.
    \item Axon -- Josephson Transmission Line (JTL).
    The transmission of the action potential originating from the Soma to adjacent neurons can be carried out by the Josephson Transmission Line (JTL). A magnetic flux quantum can move along the JTL with a short delay and low energy dissipation, and its passage through the Josephson junction is accompanied by the appearance of a spike voltage pulse on the heterostructures. This mechanism can be likened to the process observed in biological systems, where the myelin sheath covering axons enables the action potential to ``hop'' from one point to the next.
    
    \item Synapse -- Adaptive Josephson Transmission Line (AJTL).    
    Synapses exert an inhibitory or excitatory influence on the postsynaptic neuron with respect to the past activity of the pre- and post-synaptic neuron. While memory is still difficult to implement in RSFQ technology, Feldhoff et al. \cite{feldhoff2021niobium} have designed a short-term adaptation element that sets a connection weight depending on the steady state activity of the pre- and post-synaptic neuron. A common design element in RSFQ circuits is a release junction to prevent congestion. A release junction is inserted into a JTL, making it controllable by two currents. As a result, the release probability of an RSFQ from the JTL SQUID can be controlled by the current flowing through the junction. By driving the pulse sequence of the pre- and post-synaptic neuron through an L-R low-pass filter, the current through the JJ output is a function of the pulse frequency and thus of the activity level of both neurons. 
    Thermal noise adds uncertainty to the junction switching and flux release from the superconducting loop. This creates a continuous dependence of the SFQ transmission probability on the junction current balance. Recently Feldhoff et. al. \cite{feldhoff2024short} have presented an improved version of the synapse where the single junction is replaced by a two junction SQUID, called a release SQUID. This allows the critical current to be controlled by coupling an external magnetic field into the release circuit. Since the flux can be stored in another SQUID loop, the critical current can be permanently adjusted, which changes the transmission probability of SFQs passing through the synapse. This makes the synaptic connection more transparent to the connected neurons. The realization of tunable synaptic interconnections is also possible by utilizing adjustable kinetic inductance \cite{schegolev2022tunable}.
\end{enumerate}

\begin{figure}
    \centering
    \includegraphics[width=0.90\linewidth]{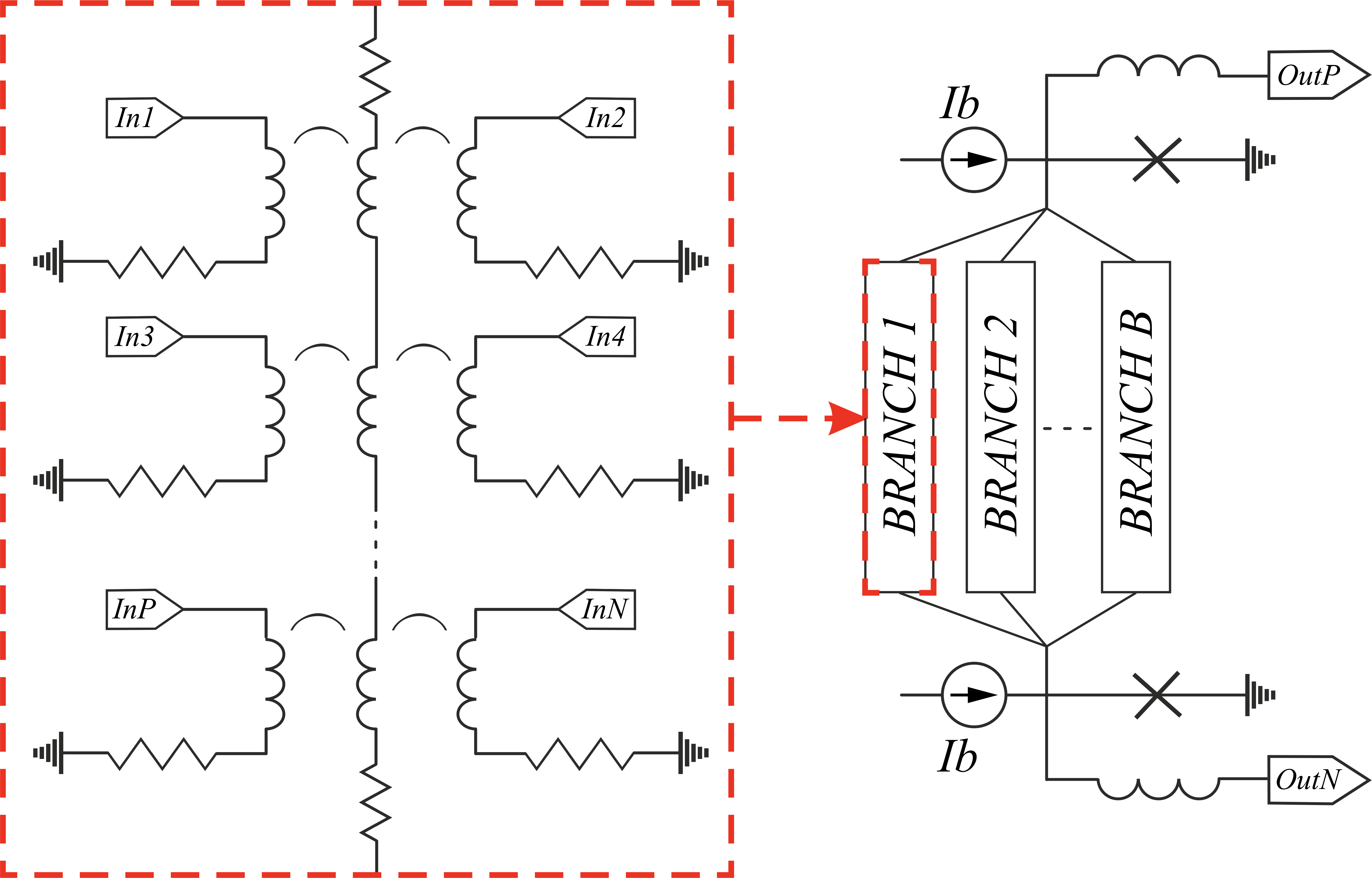}
    \caption{Schematics of an advanced neuron proposed by Karamuftuoglu et al. (right) and its input data branch structure that enables the high-fan-in feature (left)\cite{karamuftuoglu2024scalable}} 
    \label{fig:Karamuftuoglu_LIF}
\end{figure}

To enable parallel processing in a network, it is necessary to have sufficient capabilities for both input (fan-in) and output (fan-out) in the technology platform. To address the issue of limited fan-in in JJ-based neuron designs, Karamuftuoglu et al. introduced a high-fan-in superconducting neuron \cite{karamuftuoglu2024scalable}. The neuron design includes multiple branches representing dendrites, with each branch placed between two JJs that set the threshold of the neuron (see figure~\ref{fig:Karamuftuoglu_LIF}). This configuration allows for both positive and negative inductive coupling in each input data branch, supporting both excitatory and inhibitory synaptic data. The resistors on each branch create leaky behaviour in the neuron. A three hidden layer SNN using this neuron design achieved an accuracy of 97.07\% on the MNIST dataset. The network had a throughput of $8.92~GHz$ and consumed only $1.5~nJ$ per inference, including the energy required to cool the network to 4K \cite{ucpinar2024chip}.

Since memory storage is difficult to implement in RSFQ technology, magnetic Josephson junctions can be used to implement memory of past activity of pre- and post-synaptic neurons. In these devices, the wavefunction of Cooper pair extends into the ferromagnetic layer with a damped oscillatory behavior. Leveraging the physics of the interacting order parameters, Schneider et al. developed a new kind of synapse that utilizes a magnetic doped Josephson junction \cite{schneider2018tutorial}. Inserting magnetic nanoparticles into the insulating barrier between two superconductors allows for the adjustment of the Josephson junction's critical current. Since numerous particles can be placed within the same barrier, each aligned in various directions, the critical current can effectively vary across a continuous range of values, making  magnetic Josephson junctions an ideal memory element for the synaptic strength. 

\begin{figure}
    \centering
    \includegraphics[width=0.95\linewidth]{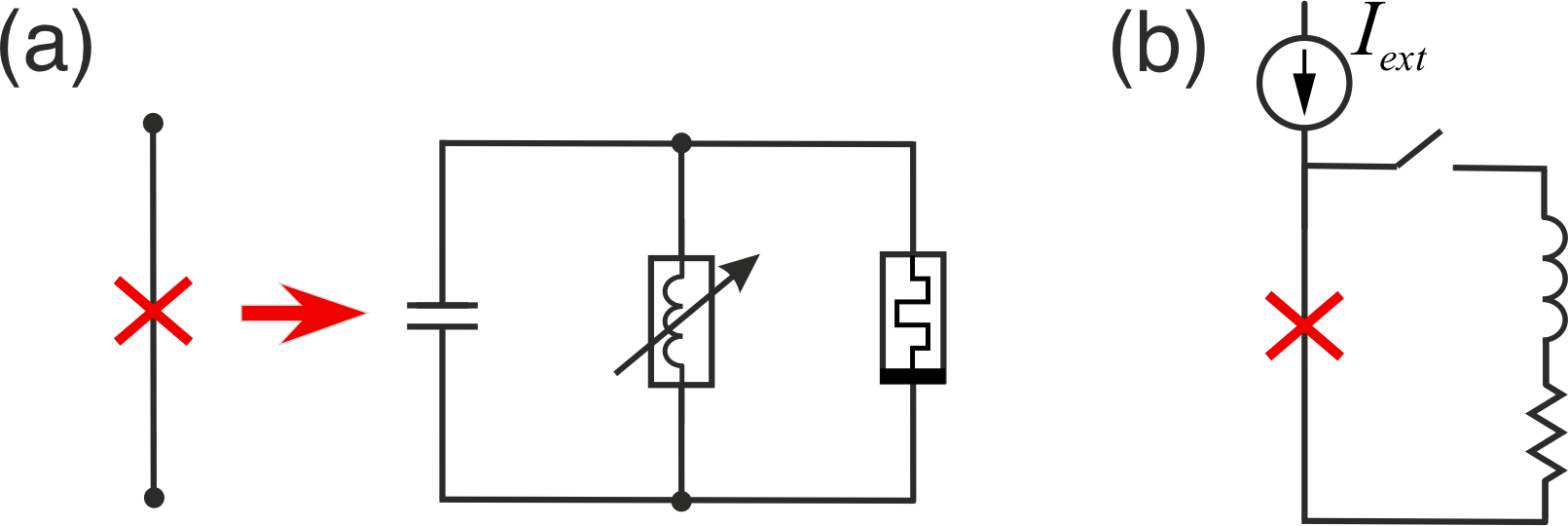}
    \caption{ (a) Equivalent circuit of a memristive Josephson junction (MRJJ) with, in order, capacitor, Josephson nonlinear device, memristor. (b) A memristive Josephson junction connected in parallel with an inductor (schematic with internal resistor, L-MRJJ) is able to mimic neuron-like bursting behaviour.
    }
    \label{fig:L_MJJ}
\end{figure}

The memristive Josephson junctions (MRJJ) can also serve as a neuron-inspired device for neuromorphic computing. The paper by Wu et al. \cite{wu2024reproduced} investigates the dynamic properties of neuron-like spiking, excitability and bursting in the memristive Josephson junction and its improved version (the inductive memristive Josephson junction, L-MRJJ). Equivalent circuits of MRJJ and L-MRJJ are shown in figures~\ref{fig:L_MJJ}a and ~\ref{fig:L_MJJ}b respectively. The MRJJ model is able to reproduce the spiking dynamics of the FitzHugh-Nagumo neuron (FHN model). Unlike the FHN model, the MRJJ model is bistable. The two class excitabilities (class $I$ and class $II$) in the Morris-Lecar neuron are reproduced by the MRJJ model based on the frequency-current curve. The L-MRJJ oscillator exhibits bursting modes analogous to the neuronal busting of the 3-D Hindmarsh-Rose (HR) model in terms of purely dynamical behaviour, but there is a discrepancy between the two models. The generating origin of the bursting patterns depends on the saddle-node and homoclinic bifurcation using a fast-slow decomposition method. The L-MRJJ model has infinite equilibria. The coupled L-MRJJ oscillators can achieve both in-phase and antiphase burst synchronisation, similar to the behaviour of coupled Hindmarsh-Rose neurons. During burst synchronisation, the L-MRJJ network is partially synchronised, but the HR network is fully synchronised. 

Adjusting neuron threshold values in spiking neural networks is important for optimizing network performance and accuracy, as this adjustment allows for fine-tuning the network's behavior to specific input patterns. Ucpinar et al. \cite{ucpinar2024chip} proposed a novel on-chip trainable neuron design, where the threshold values of the neurons can be adjusted individually for specific applications or during training.

\subsection{BrainFreeze}
Combining digital and analogue concepts in mixed-signal spiking neuromorphic architectures offers the advantages of both types of circuits while mitigating some of their disadvantages. Tschirhart et al. \cite{tschirhart2021brainfreeze} proposed a novel mixed-signal neuromorphic design based on superconducting electronics (SCE) -- BrainFreeze. This novel architecture integrates bio-inspired analogue neural circuits with established digital technology to enable scalability and programmability not achievable in other superconducting approaches. 

The digital components in BrainFreeze support time-multiplexing, programmable synapse weights and programmable neuron connections, enhancing the effectiveness of the hardware. The architecture's time-multiplexing capability enables multiple neurons within the simulated network to sequentially utilize some of the same physical components, such as the pipelined digital accumulator, thereby enhancing the hardware's effective density. Communication among neurons within BrainFreeze is facilitated through a digital network, akin to those employed in other large-scale neuromorphic frameworks. This digital network allows the sharing of wires connecting neuron cores across multiple simulated neurons, eliminating the necessity for dedicated physical wires to link each pair of neurons and significantly improves scalability. The flexible connectivity offered by the digital network also allows to implement various neural network structures by adjusting the routing tables within the network. By employing this approach, BrainFreeze leverages recent progress in SCE digital logic and insights gleaned from large-scale semiconductor neuromorphic architecture.

\begin{figure}
    \centering
    \includegraphics[width=0.90\linewidth]{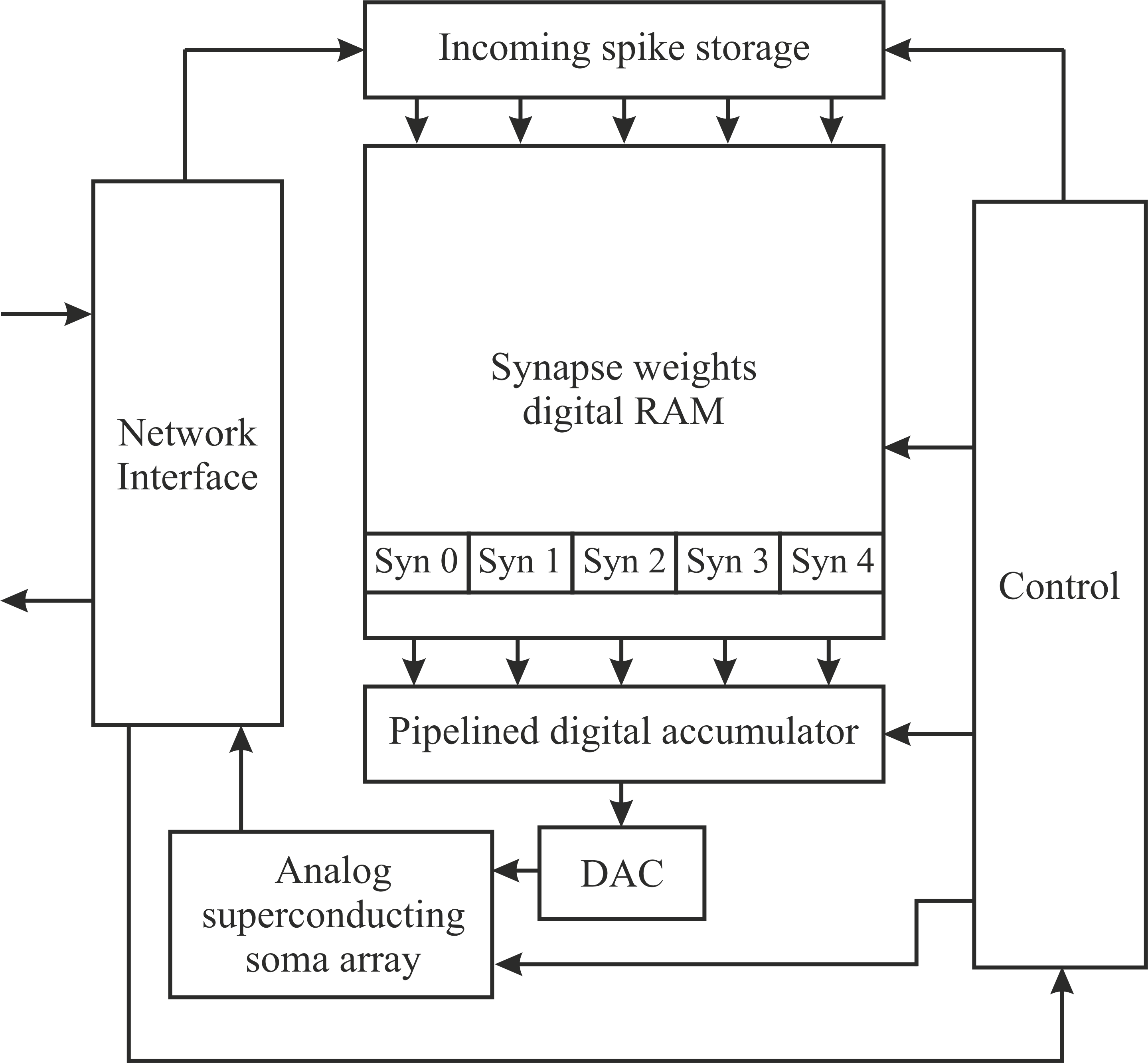}
    \caption{ A high-level block diagram of the Neuron Core as part of the BrainFreeze concept.
    }
    \label{fig:BrainFreeze_block}
\end{figure}

In its fundamental configuration, the BrainFreeze architecture consists of seven primary elements: control circuitry, a network interface, a spike buffer, a synapse weight memory, an accumulator, a digital-to-analog converter, and at least one analog soma circuit.  A schematic representation illustrating the overall architecture is provided in figure~\ref{fig:BrainFreeze_block}. The authors refer to one instance of this architecture as a Neuron Core. This architectural framework merges the scalability and programmability features allowed by superconducting digital logic with the biological suggestivity functionalities enabled by superconducting analog circuits.

\begin{figure}
    \centering
    \includegraphics[width=1.0\linewidth]{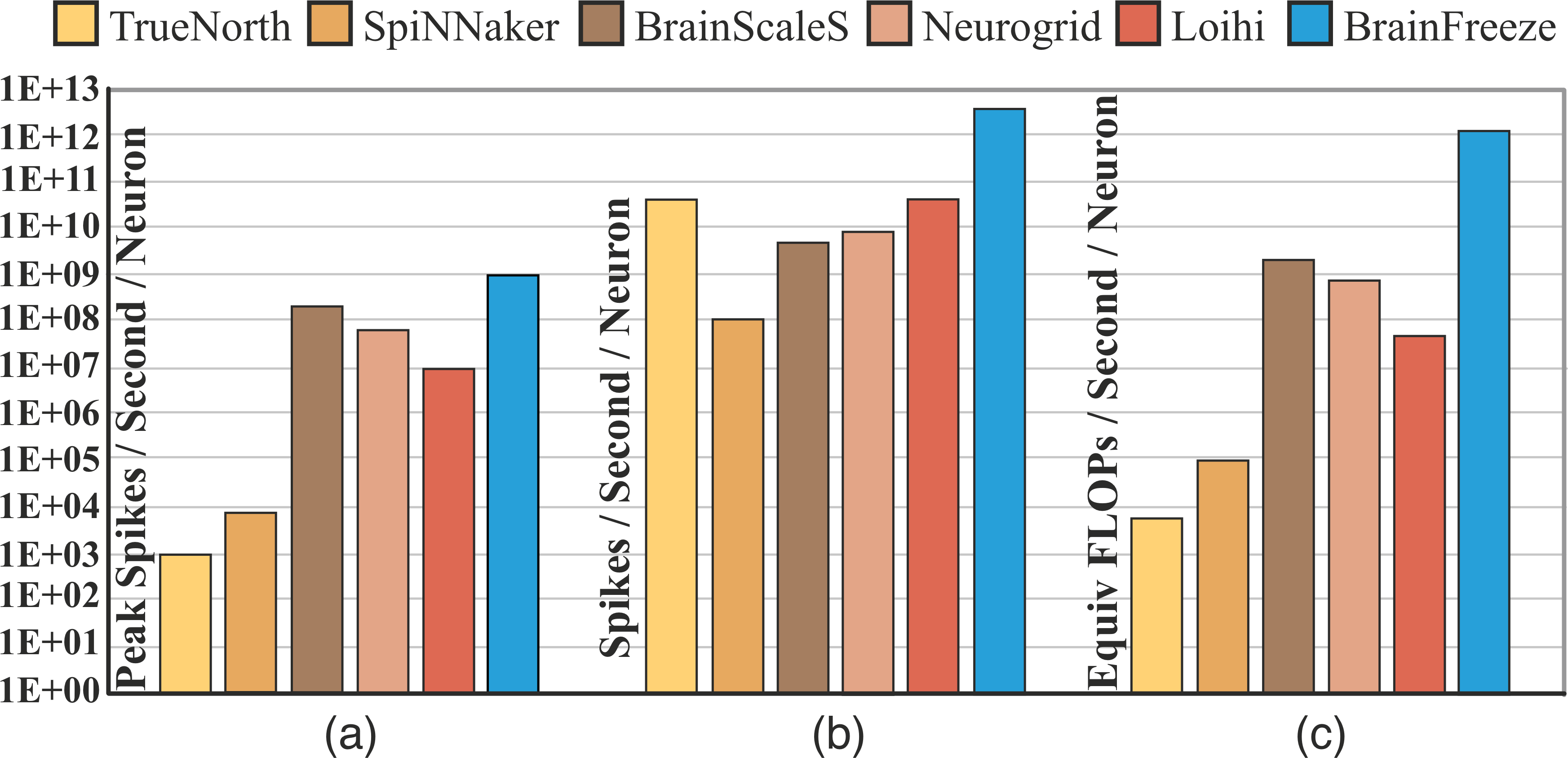}
    \caption{A comparison of state-of-the-art neuromorphic architectures with the BrainFreeze results. (a) The first comparison examines spike emission speed. (b) The second comparison examines the time and power efficiency of each architecture. (c) The final comparison evaluates the computational complexity of the neuron models embedded in each architecture. In summary, BrainFreeze shows promising potential to achieve significant improvements over existing neuromorphic approaches.
    }
    \label{fig:BrainFreezeComparison}
\end{figure}

Tschirhart et al. provided a comparison of state-of-the-art neuromorphic architectures based on CMOS such as TrueNorth, SpiNNaker, BrainScale, Neurogrid and Loihi to BrainFreeze in order to demonstrate the potential of the proposed architecture (see figure~\ref{fig:BrainFreezeComparison}). In conclusion, the findings indicate that employing a mixed-signal SCE neuromorphic approach has the potential to enhance performance in terms of speed, energy efficiency, and model intricacy compared to the current state of the art.

\subsection{Superconducting nanowire-based and phase-slip-based realisations}
For a large neuromorphic network, the number of SFQ pulses generated should be high enough to drive a large fan-out. In this case, the JJ is limited in the number of SFQ pulses it can generate. Therefore, it may be very difficult to implement a complete neuromorphic network based solely on the JJ. Schneider et al. have theoretically reported a fan-out of 1 to 10000 and a fan-in of 100 to 1 \cite{schneider2020fan}. An approximate estimate of the power dissipated for a 1-to-128 flux-based fan-out circuit for a given critical current value is reported to be 44~$aJ$. Additionally, the action potentials in JJ are not sufficiently strong to be easily detectable. An alternative to JJ could be a thin superconducting wire, also known as a superconducting nanowire (SNW). The intrinsic non-linearity exhibited by superconducting nanowires positions them as promising candidates for the hardware generation of spiking behavior. When a bias current flowing through a superconducting nanowire exceeds a threshold known as the critical current, the superconductivity breaks down and the nanowire becomes resistive, generating a voltage. The nanowire switches back to the superconducting state only when the bias current is reduced below the retrapping current and the resistive part (the ``hotspot'') cools down. Placing the nanowire in parallel with a shunt resistor initiates electrothermal feedback, resulting in relaxation oscillations \cite{toomey2018frequency}. SNWs demonstrate reliable switching from superconducting to resistive states and have shown the capability to produce a higher number of SFQ pulses as output. Toomey et al. \cite{toomey2019design, toomey2020superconducting} have proposed a nanowire-based neuron circuit that is topologically equivalent to the JJ-based bio-inspired neurons discussed above.

\begin{figure}
    \centering
    \includegraphics[width=1\linewidth]{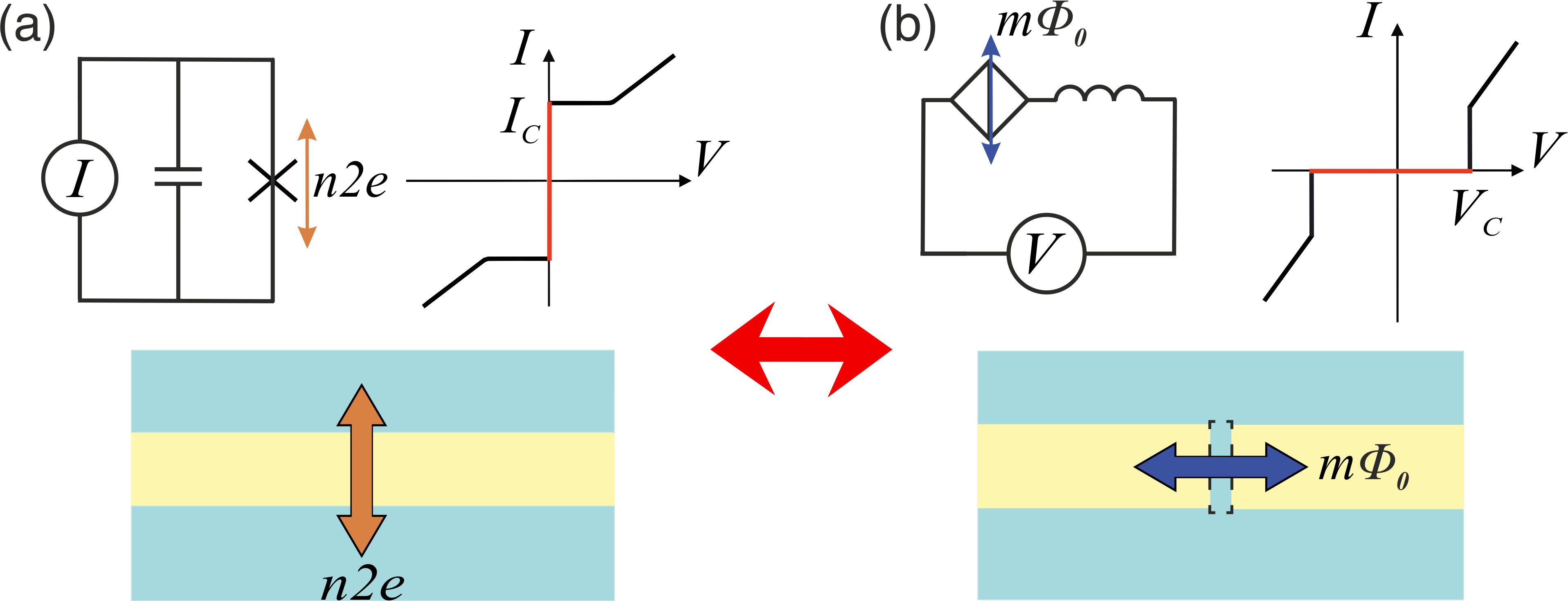}
    \caption{Illustration of the concept of charge-flux duality: 
(a) sketch of a JJ, consisting of an insulating tunnel barrier (yellow) between a superconducting island (blue)  and "ground";
(b) sketch of a QPSJ, consisting of a superconducting nanowire between an insulating island and ‘ground’. The inserts show the corresponding electrical circuits and current-voltage characteristics.
    }
    \label{fig:QPSJ_duality}
\end{figure}

Quantum phase slip can be described as the exactly dual process to the Josephson effect based on charge-flux duality (figure~\ref{fig:QPSJ_duality}). In a quantum phase slip junction (QPSJ), a magnetic flux quantum tunnels across a superconducting nanowire along with Cooper pair transport and generate a corresponding voltage across the junction \cite{astafiev2012coherent, cheng2021toward}. Unlike Josephson junctions and superconducting nanowires, QPSJs do not require a constant current bias. However, the ``flux-tunelling'' together with voltage spike generation in such systems can be used to implement neuropodic systems on a par with Josephson's. 

\begin{figure}
    \centering
    \includegraphics[width=0.50\linewidth]{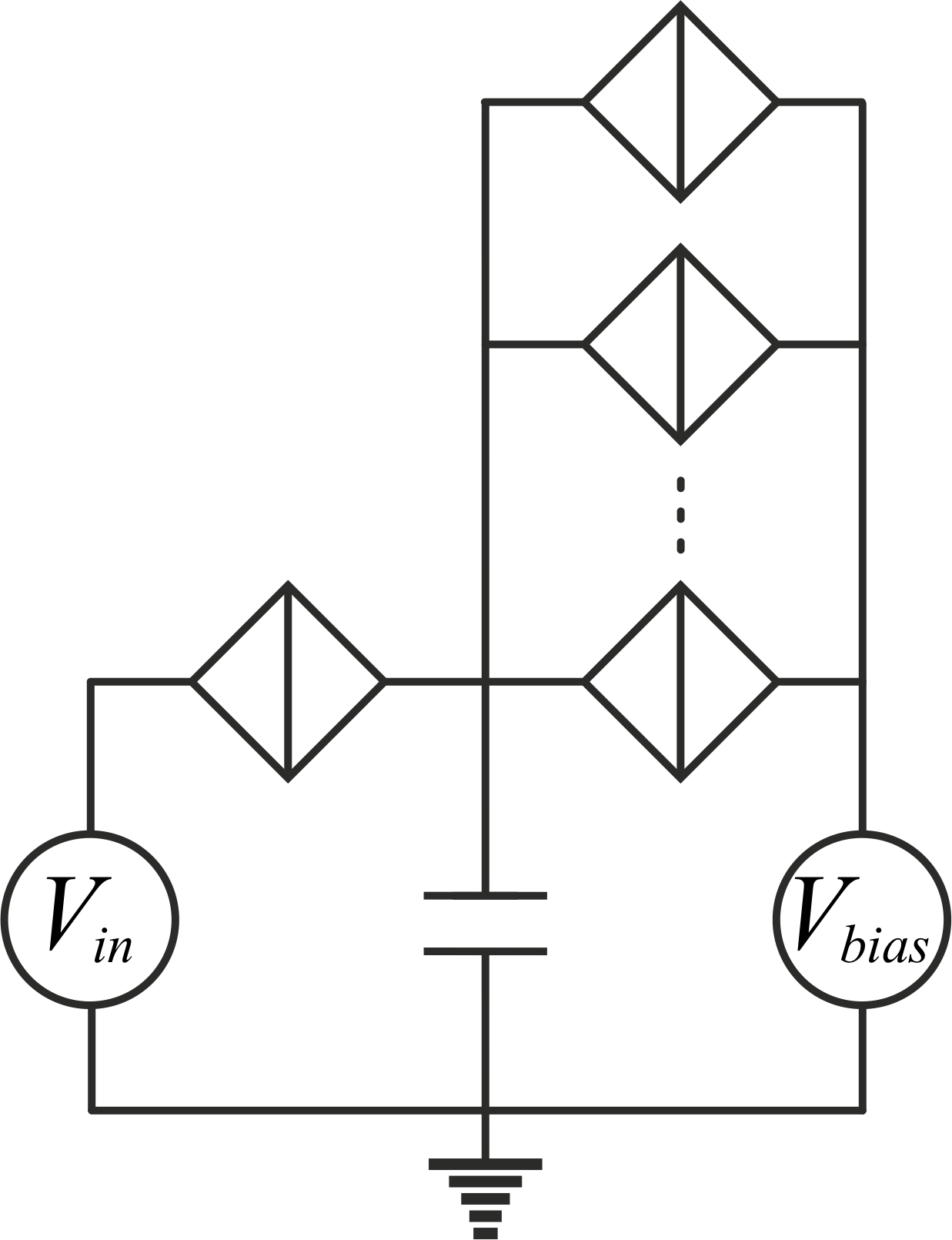}
    \caption{A QPSJ-based integrate and fire neuron circuit with integration capacitor and multiple parallel output QPS junctions \cite{cheng2018spiking}.
    }
    \label{fig:QPSJ_neuron}
\end{figure}

Cheng et al. \cite{cheng2018spiking} introduced a theoretical QPSJ-based spiking neuron design in 2018 (figure~\ref{fig:QPSJ_neuron}). When an input QPSJ fires, it charges a capacitor, building a potential that eventually exceeds the threshold of the output QPSJs connected in parallel.  A resistor can be added to maintain the same bias across all QPSJs. This behaviour is consistent with the leaky integrate-and-fire neuron model. The total ``firing'' energy for this circuit design is given by the switching energy of the QPSJ multiplied by the number of QPSJs required for ``firing'', and is estimated to be on the order of $10^{-21}~J$, compared to about $0.33~aJ$ (per switching event) for a typical JJ-based neuron. 

\section{Bio-inspired elements for optical neuromorphic systems}

Another alternative neuromorphic solution is networks with physical realisation of spikes in the form of electromagnetic field packets. The advantage of such systems as candidates for hardware implementation of optical neural networks (ONN) is also their speed of information transmission, which can be carried out at frequencies exceeding $100~GHz$ (even at room temperature!) \cite{Liu2021Research}, as well as the ability to create distributed networks with a high degree of connectivity \cite{Feldmann2021Parallel}. High bandwidth is important for practical applications such as those based on the control of hypersonic aircraft or related to the processing of radio signals. The first ONN solutions are based on silicon photonics, which is well compatible with CMOS technology, for example based on Mach-Zender interferometers \cite{Aashu2020} and micro-ring resonators \cite{Tait2019}.

A common feature of optical computing systems is the ability to perform linear operations efficiently \cite{Cheng2017Onchip, Feldmann2021Parallel}, whereas nonlinear signal transformations, including spike-based ones, pose a problem. As a result, these systems are poorly suited for implementing non-linear computations directly in the photonic domain. For this reason, hybrid solutions that combine the strengths of optical and electronic platforms are of interest. However, hybrid architectures consisting of photonic synapses and electronic spiking neurons require the use of high-speed photodetectors and analogue-to-digital converters to translate the results of optical linear computations back into the digital domain. This is complicated by high losses and packaging costs due to the need for strict alignment between lasers and waveguides.

Recently, approaches have been developed to implement photonic spiking neurons and to create fully optical spiking neural networks (OSNN) using nonlinear optical elements. Several implementations of leaky integrate-and-fire neurons have been proposed based on vertical cavity surface emitting lasers (VCSELs)\cite{Joshua2020Vertical, Zhang2021optical}, distributed feedback lasers (DFB)\cite{Hsuan2020Temporal}, phase change materials (PCM) \cite{Feldmann2019optical}. For example, VCSELs have a wide range
of laser dynamics with distinct modes corresponding to orthogonal and parallel polarisation. The excitation of a sub-ns spike in this photonic neuron is based on the injection of an external signal and the conversion of electrical impulses into optical ones, showing all the typical signs of excitability of integrate-and-fire neuronal models. At the same time, typical microwave modulation frequencies of $30-50~GHz$, combined with high efficiency, provide a sufficiently low power consumption for nonlinear conversion on the order of $10~fJ$. By creating a photonic interconnect structure (bandwidth and latency), the entire structure can, after network training, perform calculations based on an optical signal at the speed of light without additional energy cost.  This allows the dissipation in ONN using VCSEL to be decreased down to $100~aJ$ per operation. 
This looks promising from the point of view of energy efficiency \cite{Nahmias2018Neuromorphic}. However, it should be noted that most optical solutions require the use of coherent sources and detectors, as well as additional pre- and post-processing, which limits their scalability and compactness.


A detailed description of the operation of two common types of OSNN devices, based on phase change materials and semiconductor lasers, is given in the following subsections.

\subsection{Phase change materials for elements in neural networks}

Phase change materials (PCMs) are a special class of solid state materials that undergo a reversible phase transition from an amorphous to a crystalline state in response to external stimuli. Such perturbations can be both electrical and laser effects, which stimulate the release of heat in the device and, as a result, a change in temperature, causing significant changes in the optical and electrical properties of the materials \cite{Cao2020Photonic,Niloufar2017Metasurfaces}. One of the successful materials in photonic computing is the $Ge-Sb-Te$ alloy, which recently demonstrated nanosecond recording speed using optical pulses \cite{Rios2019Inmemory}. This development has led to the creation of photonic memory devices \cite{Rios2015Integrated}, switches \cite{Stegmaier2016Nonvolatile} and non-volatile computers \cite{Zheng2018GST}. In addition, PCM has established itself as a platform for neuromorphic bio-inspired on-chip computing and has already demonstrated spike-timing dependent plasticity \cite{Cheng2017chip} and control of spiking neurons \cite{Chakraborty2018Toward} in such systems.

\begin{figure}
    \centering
    \includegraphics[width=1.0\linewidth]{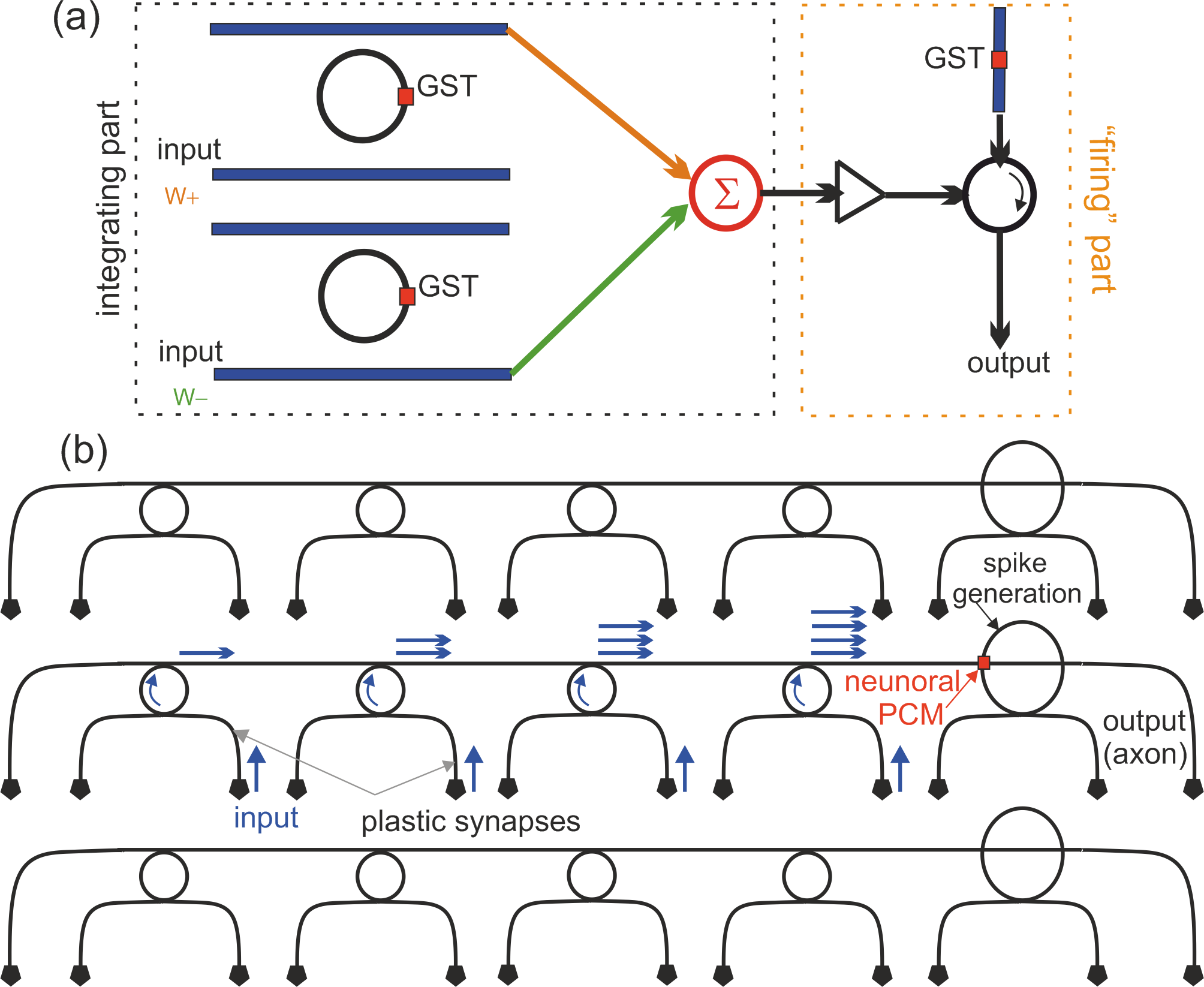}
    \caption{(a) Schematic of a bipolar integrate-and-fire neuron based on GST ring resonator devices, where the right part in the black frame is responsible for integration part, and the right part in an orange dotted frame corresponds to the ``firing'' part. (b) Schematic representation of three PCM neurons, each connected to four ring resonators (input) with different wavelengths (blue arrows). The total signal then enters the PCM cells (large rings on the right side of the figure), where the spike is generated ~\cite{Feldmann2019optical}.
    }

    \label{figPCMneuron}
\end{figure}

In \cite{Chakraborty2018Toward}, a bipolar integrate-and-fire spiking neuron is presented for the first time, including an integration unit (the black dotted block in figure~\ref{figPCMneuron}a), consisting of two ring resonators with integrated PCM based on $Ge_{2}Sb_{2}Te_{5}$ (GST) and "firing" unit (orange dotted block in figure~\ref{figPCMneuron}a). The dynamics of a spiking neuron is determined by a change in the GST phase state due to the absorption of light passing through the waveguide. 

This process causes an increase in the temperature of the material and, consequently, an increase in the amorphisation of the material, that is, the GST state can be defined as a function of the thickness of the amorphous layer in the material from the amplitude of the input pulse. The use of two ring resonators allows the neuron to receive input signals of both polarities in order to process both positive and negative weight values of $w_{\pm}$ synapses, which is an important component in information processing. In this case, the resulting amplitude of the pulse arriving at the neuron, $\Sigma$, is equal to the difference between the values of the positive and negative inputs. Thus, the integration of the membrane potential is interconnected with the amplitude of the resulting pulse arriving at the neuron, which is represented by the integrating part of the circuit in figure~\ref{figPCMneuron}а in a black dotted frame. As soon as GST reaches complete amorphisation, the membrane potential crosses its threshold. The ``firing'' action of the neuron is implemented by an additional photonic circuit (see the part highlighted by the orange dotted frame in figure~\ref{figPCMneuron}a). This circuit consists of a photonic amplifier, a circulator and a rectangular waveguide with a GST element in a crystalline state with low pulse transmission. Based on the developed bipolar neuron circuit, the possibility of recording information at subnanosecond times is shown \cite{Chakraborty2018Toward, Rios2019Inmemory}. In addition, plasticity in weighing operations of synapses \cite{Kuzum2012Nanoelectronic,Zengguang2017Onchip} were demonstrated on these PCM systems. The simplest prototype of an optical SNN was proposed, which demonstrated the scaling of individual synapses into a large-scale synaptic matrix capable of performing parallelised point computations through wavelength division multiplexing \cite{Yang2012chip}, modulating the resonant wavelength by changing the size of the waveguides. This allowed higher recording densities to be achieved and circumvented some of the problems associated with designing ring resonators whose size is comparable to the operating wavelength range. Further, in the work \cite{Chakraborty2019Photonic}, a framework was proposed to demonstrate the operation of the proposed photonic SNN platform based on ring resonators with GST in solving image classification problems.

In 2019, the first experimental implementation of the integrated photonic SNN \cite{Feldmann2019optical} was presented, consisting of three presynaptic neurons, one output (postsynaptic) neuron based on ring waveguides with GST and a six-network integrated all-optical synapses with integrated wavelength division multiplexing technology. An illustration of the experimental scheme of a neuromorphic PCM cell is shown in figure~\ref{figPCMneuron}b. The work of \cite{Feldmann2019optical} has already tested the possibility of unsupervised network learning, for which a feedback waveguide was added to the circuit design, through which part of the output neuronal pulse propagates back through the synaptic elements in the PCM. This means that the connections with all the inputs that contributed to a particular jump in the output data are strengthened, and the connections that did not contribute to the jump are weakened. As a result, the optical SNN developed was able to detect the simplest patterns.

\subsection{Spiking networks with semiconductor lasers}

Semiconductor lasers are solid-state devices whose operation is based on the properties of a semiconductor material. The power, directivity and compactness of solid-state lasers make them indispensable for high-intensity tasks, while the adaptability and efficiency of semiconductor lasers lend themselves to applications requiring precision and stability.
Fairly recently, another promising application of semiconductor lasers has appeared – neuromorphic computing. It has already been possible to observe the manifestation of bio-inspired properties of such systems, such as excitability \cite{Pei2012,Rinzel2013Nonlinear} and demonstration of nonlinear dynamics \cite{Hurtado2010Nonlinear, Takougang2012Direct}, which became promising first steps towards the implementation and study of optical spiking neurons based on semiconductor lasers.

\begin{figure}
    \centering
    \includegraphics[width=0.8\linewidth]{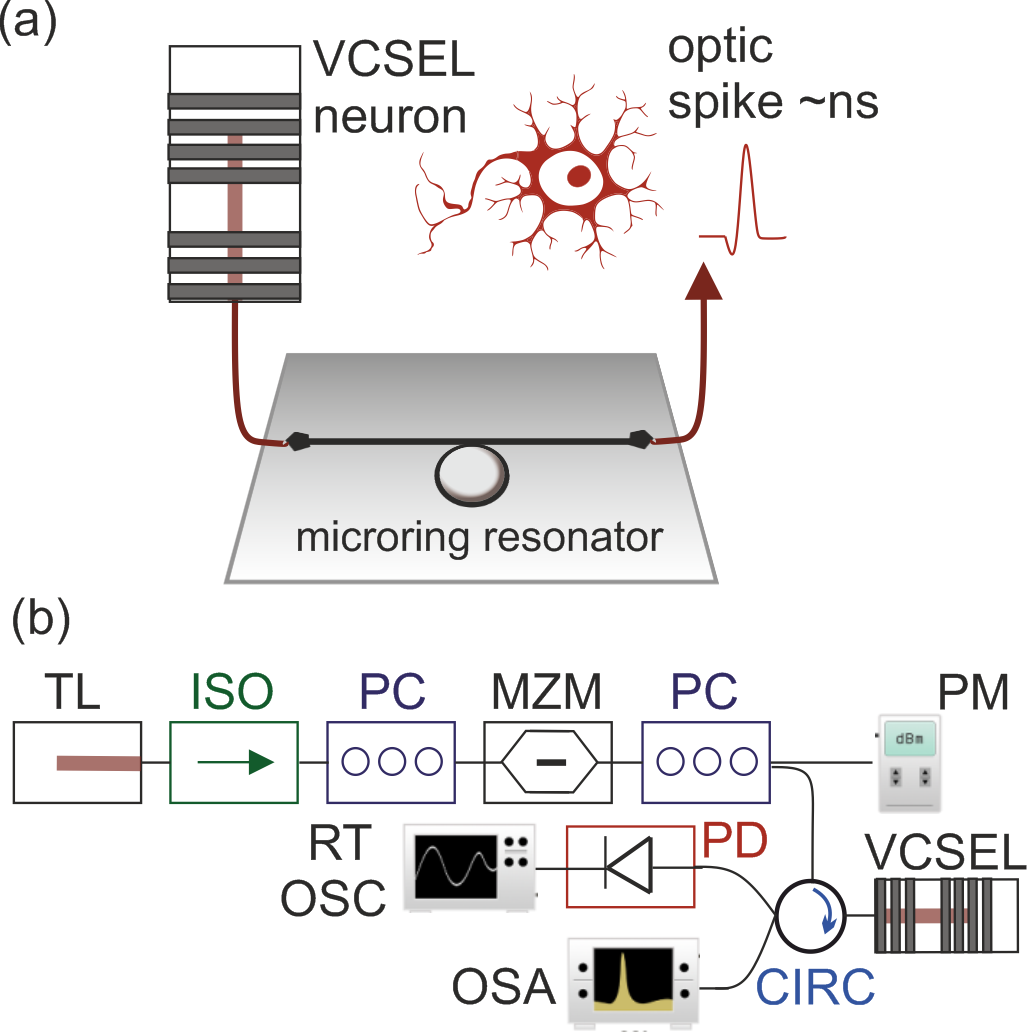}
    \caption{(a) Schematic diagram showing the basic working principle of the experimentally investigated weighting of sub-ns optical spikes produced by a VCSEL-neuron. (b) An experimental scheme of a photonic spiking neuron based on the use of the photodetector-resonant tunneling diode -- VCSEL scheme, see more details in the work \cite{Zhang2023Tunable}, where TL -- tunable laser, ISO -- optical isolator, PC -- polarization controller, MZM -- Mach–Zehnder modulator, PM -- optical power meter, CIRC -- optical circulator, OSA -- optical spectrum analyzer, PD -- photodetector, and RT OSC -- real-time oscilloscope}.
    \label{figLaserneuron}
\end{figure}

Vertical-Cavity Surface-Emitting Lasers (VCSEL) \cite{Joshua2020Vertical, Zhang2021optical} and Distributed Feedback Semiconductor Lasers (DFB-SL) \cite{Hsuan2020Temporal} are currently the main types of devices for creating and studying optical spiking neurons. The photonic elements under consideration consists of three parts: a photodetector acting as an optical-electrical converter, a receiver - a pulse converter of the micrometer range that generates pulses in response to incoming disturbances at the input, and VCSEL with a wavelength of $1550~nm$, acting as a converter of an electrical signal into an optical one. A typical scheme for implementing spiking neuron based on VCSEL is shown in figure~\ref{figLaserneuron} (a). Obtaining controlled optical spikes when exposed to exciting signals is based on the effects of polarisation switching, as well as synchronisation of phase and amplitude modulated injection locking \cite{Lu2022Frequency}.

Another way of implementing the spiking optical neuron \cite{Bruno2017Delay} in laser systems is to use a resonant tunneling diode: a semiconductor heterostructure with a double-barrier quantum well \cite{Cimbri2022Resonant}. Due to the N-shaped I-V characteristic of the diode and the effect of resonant quantum tunneling, such devices have rich nonlinear dynamics. The amplitude of optical pulses in synaptic connections can be controlled by adjusting the VCSEL bias current.
In the paper \cite{Zhang2023Tunable}, an artificial optoelectronic neuron based on an InGaAs resonant tunnel diode connected to a VCSEL has been proposed and experimentally implemented (see figure~\ref{figLaserneuron} (b)), which allows the presynaptic weight of the emitted optical pulses to be fully tuned at high speed.
The operation of the optical neuron is based on coherent injection locking of the VCSEL with a signal from an external tunable laser source (TL). The TL signal is passed through an optical isolator (ISO) to provide unidirectional communication between the lasers, and the TL light intensity is modulated using a Mach–Zehnder modulator (MZM). The polarization of the modulated signal is consistent with the orthogonally polarized VCSEL mode before it is fed into the device through an optical circulator (CIRC). Another branch of this connector is used to register  power of the VCSEL-neuron using the photodetector (PD) and the real-time oscilloscope (RT OSC) from the power meter (PM). The wavelength TL corresponds to a peak in the laser spectrum. Switching between the states (threshold and steady) is carried out by adjusting the input power. The information is encoded in the signal intensity in such a way that the ``stronger'' stimuli correspond
to a greater decrease in injection power.

The threshold characteristics and refractory period of neurons based on VCSELs \cite{Nahmias2013Leaky, Mitchell2015Optical, Paul2016Optics, Robertson2017Controlled} have already been studied. These properties naturally determine the maximum pulse response frequency and are key properties for spike processing using rate coding, photonic polarisation dynamics and excitability of neurons \cite{Hurtado2010Optical}, as well as controlled distribution spike pulses \cite{Robertson2019Electrically, Deng2017Controlled}. Furthermore, the simplest integrated photonic accelerator processor based on VCSELs has been experimentally proposed and the simplest implementation of the XOR \cite{Peng2020Temporal} classification problem has been demonstrated. It is shown that the spike laser neuron performs coincidence detection with nanosecond time resolution with a refractive period of about 0.1 ns. In addition, \cite{Xiang2019STDP} proposed a physical model for pattern recognition based on photonic STDP, which is also based on VCSELs. However, the neurons on VCSELs require a continuous power supply to maintain the behaviour of the neurons, so the advantages of energy efficiency in such systems are negated. With regard to DFB lasers, schemes have also been proposed for the implementation of passive micro-ring optical spiking neurons \cite{Ma2020Demonstration}, where neuromorphic information processing is performed, including image recognition based on STDP. Note that an important difference from VCSEL in implementing STDR on DFB-SL is the absence of wavelength conversion and the use of optical filters \cite{Xiang2019STDP,Ryan2016Photonic}.


It should be noted that spiking of optical neurons on other types of semiconductor lasers is currently being proposed. For example, ring lasers have achieved multiple spikes with a single perturbation \cite{Coomans2011Solitary}, excitation responses based on semiconductor lasers on quantum dots \cite{Kelleher2011Excitability, Kelleher2009Excitable} have been successfully modelled, threshold properties of optically injected microdisk lasers \cite{Koen2012Excitability} have been studied, and an integrated graphene-excited laser has been shown to exhibit dynamics on the order of picosecond timescales \cite{Shastri2016Spike}. However, these types of lasers are in the early stages of research for SNN compared to VCSEL and DFB-SL.

\section{Discussion and conclusion}
The opportunities for neuromorphic computing technology are immense, ranging from classical tasks such as pattern recognition and video stream analysis to central nervous system modelling and brain-computer interfaces. Challenges for the SNN include: 
neuromorphic locomotion control, neurorobotics, gait control, environmental perception, adaptation; 
computational modelling of the brain in the study of biological nervous systems (Human Brain Project); 
dynamic recognition;
prosthetics (visual and auditory implants, treatment of Parkinson's, dystonia, schizophrenia, etc.), etc.

For the time being, not many laboratories in the world are involved in these projects and tasks, as the main focus is now on the development of narrowly focused software models that solve a strictly defined problem. There is not yet a demand from large international companies or large private consumer sectors for the creation of versatile neural network models, but this moment is approaching: the problem of designing and training deep neural networks of large dimensionality is becoming more and more obvious. Nowadays, almost all neural network models are implemented in one way or another on the basis of NVIDIA's video chips, the annual production of which is very limited, and there are even certain quotas for bulk purchases. The energy required to train such networks is also a significant problem. At the same time, there is no guarantee that the neural network model is designed correctly and that the training is carried out as required: it may turn out that the accuracy of the neural network is not high enough, or there may be an overtraining effect.


The advantages of the semiconductor element base for hardware implementation of bio-inspired neural networks
are significant developments in the field of information input/output devices, digital-to-analogue and analogue-to-digital converters. 
Unfortunately, one of the main problems when trying to implement hardware parallelism of large networks is the implementation of synaptic connections. In two-dimensional parallel hardware, physical wiring only allows connections between adjacent neurons 
, whereas biological neurons are distributed in three-dimensional space and have many (thousands) connections between populations.

Memristive technologies can be instrumental in the creation of large-scale networks with a large number of synaptic connections \cite{berggren2020roadmap}.
The memristive crossbars, which are used as a kind of synaptic grid with both STDP and STDD (Spike Timing Dependent Depression).
The neurons themselves are mostly implemented using transistors. Attempts are being made to create a fully memristive neural network, but it is still not possible to realise without more conventional semiconductor devices. 

The field of superconductivity has its own distinct successes (high performance and energy efficiency) and problems (low integration density). Special attention should be paid to hybrid interdisciplinary approaches \cite{Shainline2018, Shainline2019, khan2022superconducting, shainline2023phenomenological}: for signal transmission, for light pulses transmitted from neuron to neuron via optical waveguides on a chip, for information processing and storage, and for superconducting circuits, including single-photon detectors and superconducting digital logic cells. 
Superconducting technologies make it possible to reduce the energy stored in pulses by the electromagnetic field so that a signal containing only a few photons can be used. 
To increase the compactness and efficiency of the interaction with electromagnetic radiation, current concentrating heterostructures (variable thickness bridges or Diem bridges) and multilayer thin film heterostructures can be used instead of traditional superconductor-insulator-superconductor Josephson junctions.
Finally, superconducting implementations are interesting because they allow close integration with superconducting quantum bits \cite{pashin2023bifunctional}. This offers the hope of being able to experimentally test hypotheses about the role of quantum effects in the functioning of consciousness.

In the pursuit of bio-inspiration, performance, energy efficiency, scalability and compactness, one should not fall into the extremes of the capabilities of a single element base -- it is likely that some elements of a neural network are much more convenient and efficient to implement in a different way. In this review we have tried to show not only that the field of spiking neural networks with bio-inspired properties is actively developing, but that it is developing in different directions, like an octopus with its tentacles finding the right solution, the direction of development. In our view, all the areas discussed in this article have their own strengths and weaknesses. At the moment, however, it is difficult to say which of them will ``take off''. This is why hybrid approaches are so popular at the moment, taking the advantages of their components and compensating for their mutual disadvantages.
In addition to hardware implementations and software SNNs, biological neural networks based on biological organoids and reservoir computing \cite{suarez2021learning, cai2023brain, suarez2024connectome} are on the way to creating a biocomputer \cite{smirnova2023organoid}, but we have left this area outside the scope of this review.


\begin{acknowledgments}

The concept of developing spiking neural networks was carried out with the support of the Grant of the Russian Science
Foundation No. 22-72-10075.
We are grateful to the the Foundation for the Advancement of Theoretical Physics and Mathematics ``BASIS'' (A.S. grant 22-1-3-16-1; M.B. grant 22-1-3-41-1; A.M. grant 23-2-1-16-1). 
A.S. contributed to the Introduction, the second section "CMOS-based bio-inspired neuromorphic circuits", Discussion and general edition.
M.B. contributed to the second section on memristive systems, as well as on optical neuromorphic systems.
A.M. contributed to the fourth section "Superconductor-based bio-inspired elements of neural networks".
The analysis of the possibilities of hybrid approaches was supported by the Ministry of Science and Higher Education of the Russian Federation (Agreement No. 075-15-2024-632).

\end{acknowledgments}

\bibliography{result}

\end{document}